\pdfoutput=1
\documentclass[11pt,letterpaper]{article}

\usepackage[T1]{fontenc}
\usepackage[utf8]{inputenc}
\usepackage{mathpazo}
\usepackage[left=0.92in,right=0.92in,top=0.92in,bottom=0.95in,headheight=22pt]{geometry}
\usepackage{microtype}
\usepackage{amsmath}
\usepackage{amssymb}
\usepackage{booktabs}
\usepackage{multirow}
\usepackage{array}
\usepackage{placeins}
\usepackage{xurl}
\usepackage{graphicx}
\usepackage{xcolor}
\usepackage{colortbl}
\usepackage{algorithm}
\usepackage{algpseudocode}
\usepackage[numbers,sort&compress]{natbib}
\usepackage{titlesec}
\usepackage{fancyhdr}
\usepackage{caption}

\definecolor{arxivPrimary}{HTML}{2F4E73}
\definecolor{arxivAccent}{HTML}{664865}
\definecolor{arxivMuted}{HTML}{5E6D7E}
\definecolor{sieveRow}{HTML}{EFE8F5}
\definecolor{abstractBg}{HTML}{F6F8FB}
\definecolor{abstractRule}{HTML}{D4DEE9}

\usepackage[colorlinks=true,linkcolor=arxivPrimary,citecolor=arxivPrimary,urlcolor=arxivPrimary,pdftitle={SieveVL: A Sub-Billion Dual-Granularity Evidence Reranker and Localizer for Multimodal Retrieval-Augmented Generation},pdfauthor={Haoyue Peng, Yuan Liu, Fanliang Hu, Lei Huang, Wenli Cai},pdfsubject={Computer Vision and Pattern Recognition (cs.CV); Computation and Language (cs.CL); Information Retrieval (cs.IR)},pdfkeywords={Multimodal RAG, Evidence Reranking, Visual Localization, Visual Token Pruning, Selective Abstention, Document VQA}]{hyperref}

\titleformat{\section}{\large\bfseries\color{arxivPrimary}}{\thesection}{0.6em}{}
\titleformat{\subsection}{\normalsize\bfseries\color{black!88}}{\thesubsection}{0.55em}{}
\titlespacing*{\section}{0pt}{14pt plus 3pt minus 2pt}{5pt plus 1pt}
\titlespacing*{\subsection}{0pt}{10pt plus 2pt minus 1pt}{3pt plus 1pt}
\begin{document}

\thispagestyle{plain}
\noindent{\color{arxivPrimary}\rule{\linewidth}{1.6pt}}
\vspace{8pt}
\begin{center}
    {\Large\bfseries\color{black!92} SieveVL: A Sub-Billion Dual-Granularity Evidence Reranker and Localizer for Multimodal Retrieval-Augmented Generation\par}
    \vspace{11pt}
    {\normalsize
        \textbf{Haoyue Peng}, \textbf{Yuan Liu}, \textbf{Fanliang Hu}, \textbf{Lei Huang}, \textbf{Wenli Cai}$^{*}$
    \par}
    \vspace{5pt}
    {\small\color{arxivMuted}
        Medical Artificial Intelligence Laboratory, Chongqing University Cancer Hospital,\\
        Chongqing University, Chongqing 400030, China\\[2pt]
        $^*$Corresponding author: \texttt{wenli.cai@cqu.edu.cn}
    \par}
    \vspace{5pt}
    {\footnotesize\color{arxivMuted} Preprint \;\textbar\; November 30, 2026}
\end{center}
\vspace{4pt}
\noindent{\color{abstractRule}\rule{\linewidth}{0.7pt}}
\vspace{8pt}

\noindent\fcolorbox{abstractRule}{abstractBg}{%
\begin{minipage}{\dimexpr\linewidth-2\fboxsep-2\fboxrule\relax}
\vspace{4pt}
\noindent{\small\bfseries\color{arxivPrimary}Abstract}\\[3pt]
{\small
Multimodal retrieval-augmented generation (MRAG) degrades when coarse retrievers pass visually similar distractors, redundant pages, and unpruned forms to strong readers. Existing 2B--8B rerankers perform coarse ordering only without calibration, while visual citation frameworks restrict scope to documents or couple grounding to heavy autoregressive decoding. We propose \textbf{SieveVL-0.8B}, a unified 0.873B evidence sieve (23.68M trainable parameters across unfrozen \texttt{visual.merger}, language-model LoRA, a \texttt{Yes}--\texttt{No}-initialized support head, and a bilinear pointer head) that executes candidate reranking, intra-page region localization, and hierarchical six-strip pruning in decoupled, non-autoregressive forward passes before a frozen 27B reader. Trained under a six-stratum listwise--citation objective with stripped image titles, one frozen \textbf{SieveVL-0.8B (Main)} checkpoint reaches 58.98\% 27B accuracy on MRAG-Bench ($+1.85$\,pp over CLIP, 95\% CI $[+0.07,+3.67]$; ahead of a 2.44$\times$ larger ACL~2026 canonical selector and an 8B reranker in point accuracy at 1.51$\times$--3.13$\times$ lower latency, reversing CLIP's no-gold distraction drop by $+3.02$\,pp) and outperforms the 2B baseline on 5/6 metrics across all 374 Visual-RAG~\S5.1 pools (62.57\% Hit@5; 19.09\% Recall@10), cuts 50.9\%--84.5\% of reader visual tokens across XFUND, MP-DocVQA, and zero-shot English FUNSD given OCR/layout boxes (96.43\% Band Hit@1; 0.0\% wrong-field recovery), and attains 0.8755 ANLS and 81.83\% isolated answer-page accuracy on the 5,019-question ICDAR RRC MP-DocVQA hidden test server (2026-11-26 snapshot, numerically placing~\#5). Furthermore, its null-anchored listwise calibration variant (\texttt{Null-Cal}) enables zero-tuning $\tau{=}0$ abstention on MRAG-Bench (57.58\% accuracy, $-12.5\%$ visual tokens) and reaches 0.8714 Top-2 ANLS on MP-DocVQA \texttt{val}.
}
\vspace{4pt}
\end{minipage}%
}
\vspace{6pt}

\noindent{\footnotesize\textbf{\color{arxivPrimary}Keywords:} Multimodal retrieval-augmented generation $\cdot$ Evidence reranking $\cdot$ Fine-grained visual localization $\cdot$ Visual token pruning $\cdot$ Selective abstention $\cdot$ Document VQA}
\vspace{6pt}

\section{Introduction}
\label{sec:intro}

Retrieval-augmented generation (RAG)~\cite{lewis2020rag} mitigates parametric knowledge gaps by conditioning language models on retrieved evidence. Extending RAG to natural images, tables, and visually rich documents---multimodal RAG (MRAG)~\cite{chen2022murag,hu2025mragbench,ali2026pythiarag,zhang2026acra}---exposes a dual-granularity mismatch between coarse retrieval and downstream reading. At the \emph{candidate level}, contrastive dual encoders such as CLIP~\cite{radford2021clip} rank items by global visual similarity rather than answer utility, frequently admitting wrong-entity distractors. On MRAG-Bench~\cite{hu2025mragbench}, 597 of 1,353 test queries have no gold image in the retrieved CLIP Top-5 pool; passing CLIP Top-3 images to a frozen 27B reader (\texttt{qwen3.8-27b-fp8}) lowers accuracy on these queries from 50.08\% (query-only) to 48.07\% ($-2.01$\,pp). At the \emph{intra-document level}, unpruned multi-page documents and dense scanned forms inflate visual tokens and distract strong readers: on 4--10 page MP-DocVQA~\cite{tito2023mpdocvqa} documents, feeding all pages consumes 4,700.2 visual tokens per query and drops 27B ANLS from 0.9490 (gold page) to 0.8589 ($-9.01$\,pp).

Existing peer-reviewed methods address these two bottlenecks in isolation. Multimodal rerankers, including LamRA~\cite{liu2025lamra}, EviRank~\cite{du2026evirank}, miniReranker~\cite{fan2026minireranker}, and training-free Utility-VES~\cite{luo2026utility}, improve coarse ordering with 2B--8B vision--language models (VLMs), yet they neither pinpoint answer-bearing sub-regions inside dense pages nor calibrate unnormalized scores against cross-query drift for abstention. Conversely, visual-document and grounding frameworks---such as RegionRAG~\cite{li2026regionrag}, iterative RL agents (VRAG-RL~\cite{wang2025vragrl}), and autoregressive citation models (MCite-RL~\cite{zhao2026mciterl}, Doc-CoB~\cite{mo2026doccob})---either specialize solely to documents, incur multi-turn latency, or couple grounding to heavy autoregressive decoding, sacrificing the modularity and speed needed to screen heterogeneous pools before a strong reader.

This paper asks whether a \emph{single sub-billion} VLM can unify both granularities in one decoupled, non-autoregressive pass. We present \textbf{SieveVL-0.8B}, built on \texttt{Qwen3.5-0.8B} (0.873B parameters; 23.68M trainable across unfrozen \texttt{visual.merger} layers, language-model LoRA~\cite{hu2022lora}, and two lightweight task heads). In a single forward pass over a query and candidate image, page, table, or region-indexed form, the final token state feeds a \texttt{Yes}--\texttt{No}-initialized support head for item/page reranking and selective abstention, while dedicated unit-marker states interact with it through a bilinear pointer head to score fine-grained regions without rewriting source text. For dense full-page forms exceeding single-window length, the same frozen checkpoint runs a layout-agnostic hierarchical six-strip pass to select the evidence band and optionally refine it into a tight crop. To stabilize abstention across empty pools, we further introduce a null-anchored listwise calibration extension (\texttt{Null-Cal}) anchored at zero.

We evaluate one frozen checkpoint across five peer-reviewed benchmarks under identical candidate pools, same-GPU timing (NVIDIA A100 MIG \texttt{3g.40gb}), and a shared frozen 27B reader, directly comparing against reproduced ACL~2026 Utility-VES (2.13B) and Qwen3-VL-Reranker-8B (8.77B). Our main contributions are fourfold:
\begin{enumerate}
    \item \textbf{Unified sub-billion dual-granularity design.} SieveVL-0.8B unifies candidate/page reranking and non-generative intra-page pointer localization in one 0.873B forward pass, trained with a balanced six-stratum listwise--BCE--citation objective and stripped image titles for pixel-grounded discrimination.
    \item \textbf{Pareto-superior natural-image and multi-page reranking.} At 1.46$\times$--3.57$\times$ lower latency than 2B--8B baselines, SieveVL-0.8B reaches 58.98\% MRAG-Bench 27B accuracy ($+1.85$\,pp over CLIP, 95\% CI $[+0.07, +3.67]$; beating canonical ACL~2026 \texttt{true\_logit} at 58.31\% and 8B at 57.65\%, while statistically tying \texttt{p\_true} at 59.20\%) and surpasses ACL~2026 Utility-VES on five of six metrics across all 374 Visual-RAG~\S5.1 pools (62.57\% Hit@5; 19.09\% Recall@10). On zero-shot MP-DocVQA, it beats Utility-VES in Page Hit@1 (76.47\% vs.\ 74.79\%) and attains 0.8755 ANLS and 81.83\% isolated page accuracy on the 5,019-question ICDAR RRC hidden test server (2026-11-26 snapshot, numerically placing \#5).
    \item \textbf{Zero-shot cross-lingual localization and visual-token pruning.} Trained solely on Chinese form regions, SieveVL-0.8B transfers zero-shot to English FUNSD forms given OCR/layout entity boxes (0.9461 panel Citation F1; 96.43\% full-page Band Hit@1 across 42 eligible pages) and cuts 50.9\%--84.5\% of real 27B reader visual tokens across XFUND, MP-DocVQA, and full-page FUNSD, while counterfactual wrong-field cropping yields 0.0\% accuracy ($>$94\% abstention).
    \item \textbf{Null-anchored calibration and empirical scope boundaries.} Our \texttt{Null-Cal} objective enables zero-tuning $\tau{=}0$ abstention ($-12.5\%$ visual tokens on MRAG-Bench) while improving document cropping and Top-2 ANLS; our ablations further characterize sub-billion cross-query AUROC saturation alongside negative distillation and cross-corpus hard-negative findings.
\end{enumerate}
The remainder of this paper is organized as follows. Section~\ref{sec:related} reviews related work and positions SieveVL-0.8B against 2025--2026 peer-reviewed systems. Section~\ref{sec:method} details the architecture, objectives, and hierarchical full-page inference. Sections~\ref{sec:setup} and~\ref{sec:results} present the experimental protocols and results across all five benchmarks, and Section~\ref{sec:conclusion} discusses limitations and concludes.

\section{Related Work}
\label{sec:related}

\subsection{Multimodal Retrieval-Augmented Generation}
Early multimodal QA datasets such as MultiModalQA~\cite{talmor2021multimodalqa} and WebQA~\cite{chang2022webqa} established multi-hop reasoning over mixed text, tables, and images, followed by open-domain generative architectures like MuRAG~\cite{chen2022murag}. Recent knowledge-guided multimodal retrieval research has emphasized structured knowledge aggregation, cross-modal retrieval efficiency, and answer robustness: Pythia-RAG~\cite{ali2026pythiarag} builds a unified multimodal knowledge graph for QA, Flow-RAG~\cite{zhang2026flowrag} introduces gated flow propagation over knowledge graphs, ACRA~\cite{zhang2026acra} designs an adaptive chain retrieval architecture for knowledge-augmented VQA, and Yang et al.~\cite{yang2026evidencechain} propose evidence-chain-driven lightweight retrieval to improve precision and cost over LLM-guided selection. Complementary studies address bilingual hierarchical knowledge linking in HiLINK~\cite{jeong2025hilink}, visual QA robustness under challenging inputs in R-VQA~\cite{chowdhury2025rvqa}, and fast cross-modal indexing via online prototype hashing in LPGOH~\cite{peng2025lpgoh}. On the evaluation side, vision-centric benchmarks have exposed sharp retrieval-side failure modes: MRAG-Bench~\cite{hu2025mragbench} demonstrates that visual evidence helps only when fine-grained visual perspectives match the query, and Visual-RAG~\cite{wu2026visualrag} highlights severe retrieval degradation across dense same-species pools of 200--300 visually similar images. Our study evaluates directly on these peer-reviewed benchmarks under realistic fixed-retriever pools without leaking ground-truth images into candidate sets.

\subsection{Multimodal Evidence Reranking and Utility Selection}
To bridge coarse bi-encoders~\cite{radford2021clip} and late-interaction multimodal retrievers~\cite{lin2024preflmr,faysse2025colpali} with expensive generators, several 2025--2026 peer-reviewed multimodal rerankers adapt large vision--language models. LamRA~\cite{liu2025lamra} inserts LoRA adapters into 7B--8B LMMs for universal retrieval and reranking, while Progressive Re-ranking~\cite{min2026progressive} applies multi-stage curriculum learning for MRAG ordering. EviRank~\cite{du2026evirank} recasts image reranking as structured constraint verification via typed evidence frames, and miniReranker~\cite{fan2026minireranker} accelerates multimodal reranking through inference-time visual cache reuse and interaction sparsity. Region-R1~\cite{hu2026regionr1} applies reinforcement learning to learn \emph{query-side} region cropping before reranking, rather than localizing answer-bearing evidence on candidate documents. Most closely related to our coarse-ranking stage, Utility-VES~\cite{luo2026utility} formulates visual evidence selection as latent answer-space information gain, using a training-free 2.1B--2.6B surrogate VLM (\texttt{Qwen3-VL-2B-Instruct}) to score candidates by first-step \texttt{True}-token logits on MRAG-Bench and Visual-RAG. SieveVL-0.8B differs in three concrete respects: (i) it trains a native sub-billion (0.873B) scorer that is 2.44$\times$ smaller and 1.46$\times$--1.65$\times$ faster than 2B surrogates without relying on inference-only cache approximations; (ii) it unifies candidate/page reranking with \emph{candidate-side} intra-page region pointer localization in the same non-autoregressive pass, a capability absent in LamRA, Progressive Re-ranking, EviRank, miniReranker, Region-R1, and Utility-VES; and (iii) we benchmark Utility-VES directly under identical natural CLIP/embedding pools and a shared 27B reader.

\subsection{Visual Document Retrieval, Grounded Citations, and Context Pruning}
Layout-aware document pre-training such as LayoutLMv3~\cite{huang2022layoutlmv3}, OCR-free visual document retrieval~\cite{faysse2025colpali}, and multi-page DocVQA~\cite{tito2023mpdocvqa} show that raw page images preserve layout cues lost in text-only pipelines, yet incur steep visual-token costs when multiple high-resolution pages enter an MLLM reader. Recent peer-reviewed document RAG pipelines tackle this trade-off along three trajectories. First, page-selection and document-level systems such as AVIR~\cite{li2025avir} perform adaptive visual in-document retrieval over a frozen LVLM (e.g., 7B) to prune irrelevant pages on MP-DocVQA, while VDocRAG~\cite{tanaka2025vdocrag}, mPLUG-DocOwl2~\cite{hu2025docowl2}, Docopilot~\cite{duan2025docopilot}, and DocVLM~\cite{nacson2025docvlm} compress or encode multi-page contexts inside dedicated multi-billion document models; these remain page-level or document-specific rather than unified across natural-image MRAG and sub-page form grounding. Second, region-level and agentic document systems---including RegionRAG~\cite{li2026regionrag} for patch-grouped visual document retrieval, ViDoRAG~\cite{wang2025vidorag} for dynamic multi-agent iterative reasoning, and VRAG-RL~\cite{wang2025vragrl} for reinforcement-learned multi-turn visual perception---either restrict scope to document retrieval or incur multi-turn iterative calls. Third, grounded generation frameworks such as MCite-RL~\cite{zhao2026mciterl} and visual chain-of-boxes reasoning in Doc-CoB~\cite{mo2026doccob} align visual citations with generated answers, but tightly couple grounding to heavy autoregressive decoding. Form understanding benchmarks---FUNSD~\cite{jaume2019funsd} in English and XFUND~\cite{xu2022xfund} across seven additional languages---provide human-verified key--value bounding boxes ideal for verifying spatial grounding fidelity. By contrast, SieveVL-0.8B decouples localization from generation: a single 0.873B front-end selects natural images, document pages, layout strips, or tight entity crops in one non-autoregressive pass, cutting real visual tokens by 50.9\%--84.5\% before any frozen strong reader executes.

\subsection{Selective Prediction and Abstention in RAG}
Adaptive retrieval frameworks such as Self-RAG~\cite{asai2024selfrag} train generators to emit reflection tokens controlling whether retrieved context should be used. In decoupled MRAG pipelines with frozen strong readers, abstention must instead rely on the external reranker's cross-query confidence calibration, a classic challenge in modern neural classifiers~\cite{guo2017calibration}. Our evaluation systematically compares gold-in-pool AUROC, query-grouped 5-fold cross-validated abstention, and a null-anchored listwise objective across 0.87B, 2.13B, and 8.77B models, revealing both practical token savings and the capacity boundary of sub-billion cross-query calibration. Table~\ref{tab:positioning} positions SieveVL-0.8B against the closest 2025--2026 peer-reviewed multimodal reranking, document retrieval, and visual grounding systems.

\begin{table}[tbp]
\centering
\scriptsize
\setlength{\tabcolsep}{2.5pt}
\begin{tabular*}{\linewidth}{@{\extracolsep{\fill}}llccccc@{}}
\toprule
\textbf{System (venue)} & \textbf{Scale / paradigm} & \shortstack{\textbf{Single-pass}\\\textbf{non-AR}} & \shortstack{\textbf{Item/page}\\\textbf{rerank}} & \shortstack{\textbf{Intra-page}\\\textbf{localize}} & \shortstack{\textbf{Decoupled}\\\textbf{reader}} & \shortstack{\textbf{Zero-tune}\\\textbf{abstain}} \\
\midrule
LamRA~\cite{liu2025lamra} (CVPR'25) & 7B--8B LMM & Yes & Yes & No & Yes & No \\
AVIR~\cite{li2025avir} (MMAsia'25) & Frozen LVLM (7B) & Adaptive & Pages only & No & Yes & No \\
ViDoRAG~\cite{wang2025vidorag}, VRAG-RL~\cite{wang2025vragrl} ('25) & Agent / RL & No (turns) & Docs only & Iterative & No & No \\
RegionRAG~\cite{li2026regionrag} (AAAI'26) & Region VLM & Yes & Docs only & Doc regions & Yes & No \\
Utility-VES~\cite{luo2026utility} (ACL'26) & 2.1B Free & Yes & Yes & No & Yes & No \\
Region-R1~\cite{hu2026regionr1} (ACL'26 Find.) & RL Crop VLM & No (RL) & Yes & Query-side & Yes & No \\
Progressive~\cite{min2026progressive} (ACL'26 Find.) & Curriculum VLM & No (2-stage) & Yes & No & Yes & No \\
miniReranker~\cite{fan2026minireranker} (EMNLP'26 Find.) & Sparse VLM & Yes & Yes & No & Yes & No \\
EviRank~\cite{du2026evirank} (EMNLP'26) & Evidence VLM & No (gen.) & Yes & No & Yes & No \\
MCite-RL~\cite{zhao2026mciterl}, Doc-CoB~\cite{mo2026doccob} ('26) & Citation / CoB & No (AR) & No & AR-coupled & No & No \\
\midrule
\rowcolor{sieveRow}
\textbf{SieveVL-0.8B (Ours)} & \textbf{0.873B (23.7M)} & \textbf{Yes} & \textbf{Yes} & \textbf{Yes (Ptr)} & \textbf{Yes} & \textbf{Yes (Null)} \\
\bottomrule
\end{tabular*}
\caption{Capability positioning of \textbf{SieveVL-0.8B} relative to closely related 2025--2026 peer-reviewed multimodal reranking, visual document retrieval, and grounded citation systems (``Find.'' denotes ACL/EMNLP Findings). SieveVL-0.8B uniquely combines sub-billion scale, single-pass non-autoregressive scoring, dual-granularity candidate reranking and intra-page localization, reader decoupling, and zero-tuning abstention.}
\label{tab:positioning}
\end{table}

\section{Methodology}
\label{sec:method}

\subsection{Dual-Granularity Problem Formulation}
Given a multimodal query $q$ (optionally accompanied by a query image $I_q$, as in MRAG-Bench) and a candidate pool $\mathcal{C} = \{c_1, \dots, c_M\}$ returned by a coarse retriever or obtained from pages/layout strips of a document, each candidate $c_i$ may optionally expose an ordered list of $K_i$ indexable sub-units $\mathcal{U}_i = \{u_{i,1}, \dots, u_{i,K_i}\}$ (such as table cells or document bounding boxes). A dual-granularity evidence sieve must produce two coupled outputs in a single forward pass:
\begin{equation}
    f_\theta(q, c_i) \mapsto \Big(s_i \in \mathbb{R},\; \mathbf{p}_i = (p_{i,1}, \dots, p_{i,K_i}) \in [0, 1]^{K_i}\Big),
\end{equation}
where $s_i$ is the candidate-level support logit indicating whether $c_i$ provides useful evidence for $q$, and $p_{i,k}$ is the probability that sub-unit $u_{i,k}$ supports the answer. At inference time, top-$k$ candidates ranked by $s_i$ are retained (or abstained when $\max_i s_i < \tau$), and for region-indexed documents, units with $p_{i,k} \ge \gamma$ are mapped deterministically back to source pixel coordinates to form a tight crop passed to a frozen downstream reader $\mathcal{R}$. Crucially, SieveVL-0.8B never generates or paraphrases citation text; localization selects input indices directly.

\subsection{Unified Single-Pass Architecture}
\begin{figure*}[t]
    \centering
    \includegraphics[width=\textwidth]{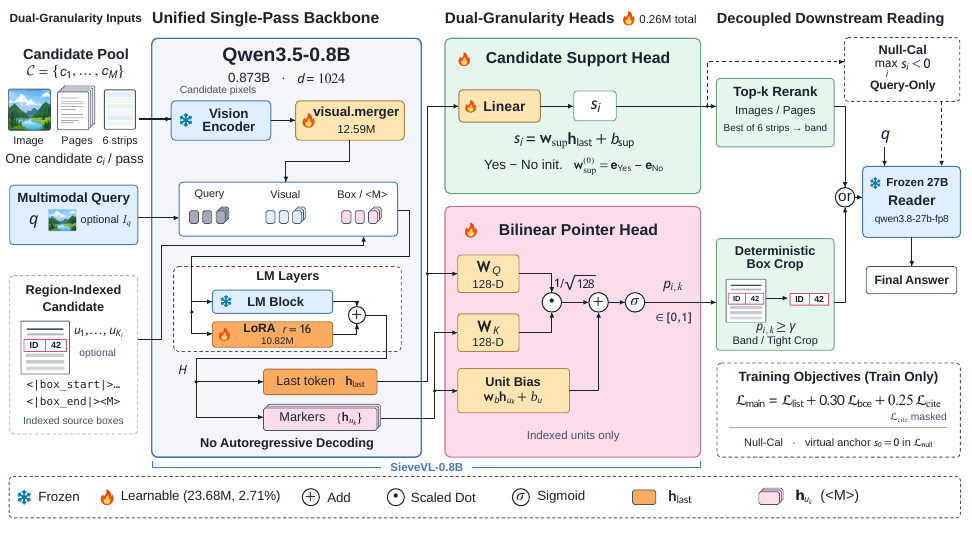}
    \caption{Overview of \textbf{SieveVL-0.8B}. A single non-autoregressive forward pass through \texttt{Qwen3.5-0.8B} (frozen vision encoder, unfrozen \texttt{visual.merger}, and language-model LoRA) extracts $h_{\mathrm{last}}$ for the \texttt{Yes}--\texttt{No}-initialized support head and marker states $\{h_{u_k}\}$ for the bilinear pointer head, pruning images, pages, bands, or tight regions before the decoupled 27B reader.}
    \label{fig:arch}
\end{figure*}

Figure~\ref{fig:arch} illustrates the architecture of SieveVL-0.8B. We build on the post-trained native vision--language backbone \texttt{Qwen3.5-0.8B} ($873{,}438{,}784$ total parameters, hidden size $d{=}1024$) and remove autoregressive decoding entirely during scoring.

\paragraph{Modality-Aware Serialization without Shortcut Leakage}
All inputs share one concise prompt prefix (\emph{``Decide whether the current candidate provides useful evidence for the question. Answer only `Yes' or `No'.''}) followed by the question and candidate payload. For natural images and document pages, raw pixels are fed directly with \textbf{all image titles stripped} during training so the model cannot exploit lexical title shortcuts instead of visual evidence. For region-indexed documents of size $(W, H)$, each candidate unit box $(x_1, y_1, x_2, y_2)$ is normalized to $[0, 1000]$ coordinates and serialized as \texttt{<|box\_start|>(a,b),(c,d)<|box\_end|><M>}, where \texttt{<M>} is a single dedicated marker token whose position indexes $u_k$ (unit order is randomly permuted per training step to eliminate positional bias). Tables serialize cells as \texttt{header: value<M>}.

\paragraph{Selective Parameter Adaptation}
Freezing the vision encoder while adapting only language-model projections limits fine-grained visual discrimination (confirmed in our Round-1 ablation). We therefore jointly optimize three parameter groups totaling $23{,}675{,}394$ trainable parameters (2.71\% of the backbone): (i) \textbf{unfrozen \texttt{visual.merger}} ($12{,}588{,}544$ parameters, learning rate $5{\times}10^{-5}$) to align visual patch tokens with evidence-relevance semantics; (ii) \textbf{LoRA adapters}~\cite{hu2022lora} ($r{=}16$, $\alpha{=}32$, dropout $0.05$, learning rate $1.5{\times}10^{-4}$; $10{,}822{,}656$ parameters) across all language-model self-attention, linear-attention, and MLP projections; and (iii) \textbf{dual scoring heads} ($264{,}194$ parameters, learning rate $2{\times}10^{-4}$).

\paragraph{Support Head with \texttt{Yes}--\texttt{No} Initialization}
Let $\mathbf{h}_{\mathrm{last}} \in \mathbb{R}^{d}$ denote the final-layer hidden state at the last non-padding token. Rather than random initialization, we initialize the support linear projection $\mathbf{w}_{\mathrm{sup}} \in \mathbb{R}^{1 \times d}$ from the difference between the base language-modeling head's \texttt{Yes} and \texttt{No} token embeddings:
\begin{equation}
    \mathbf{w}_{\mathrm{sup}}^{(0)} = \mathbf{e}_{\mathrm{Yes}} - \mathbf{e}_{\mathrm{No}}, \qquad b_{\mathrm{sup}}^{(0)} = 0, \qquad s(q, c) = \mathbf{w}_{\mathrm{sup}} \mathbf{h}_{\mathrm{last}} + b_{\mathrm{sup}}.
\end{equation}
This initialization preserves zero-shot discriminative geometry from step 0 (yielding 91.93\% Gold@3 on MRAG-Bench even before fine-tuning) and accelerates convergence within a single epoch.

\paragraph{Bilinear Pointer Head for Region Grounding}
For region- or cell-indexed candidates, let $\mathbf{h}_{u_k} \in \mathbb{R}^{d}$ be the hidden state at the $k$-th marker token \texttt{<M>} from the \emph{same} forward pass. A query-conditioned bilinear scorer with learned projections $\mathbf{W}_Q, \mathbf{W}_K \in \mathbb{R}^{128 \times d}$ and unit bias $\mathbf{w}_b \in \mathbb{R}^{1 \times d}$ computes per-unit evidence logits:
\begin{equation}
    z_k = \frac{\big(\mathbf{W}_Q \mathbf{h}_{\mathrm{last}}\big)^\top \big(\mathbf{W}_K \mathbf{h}_{u_k}\big)}{\sqrt{128}} + \mathbf{w}_b \mathbf{h}_{u_k} + b_u, \qquad p_k = \sigma(z_k).
\end{equation}
Conditioning on $\mathbf{h}_{\mathrm{last}}$ ensures that changing the question over an identical document image immediately shifts attention to the newly queried field, which we validate via paired same-panel question-switching and counterfactual cropping.

\subsection{Grouped Multi-Granularity Training Objectives}
We organize training instances into query-grouped mini-batches $\mathcal{G}$, where each group $g = (q, \mathcal{C}_g)$ with $\mathcal{C}_g = \{(c_j, y_j, \mathbf{r}_j)\}_{j=1}^{m}$ pairs candidates sharing query $q$, with candidate support label $y_j \in \{0, 1\}$ and gold unit membership $\mathbf{r}_j \in \{0,1\}^{K_j}$ whenever fine-grained annotations exist.

\paragraph{Main Objective (\texttt{SieveVL-0.8B})}
Within each group containing a verified positive candidate $p$ alongside native distractors, we optimize a joint loss combining listwise InfoNCE~\cite{cao2007listnet,oord2018infonce}, candidate-level binary cross-entropy, and granularity-masked region citation loss:
\begin{equation}
    \mathcal{L}_{\mathrm{main}} = \mathcal{L}_{\mathrm{list}} + \lambda_{\mathrm{bce}} \mathcal{L}_{\mathrm{bce}} + \lambda_{\mathrm{cite}} \mathcal{L}_{\mathrm{cite}},
\end{equation}
where $\lambda_{\mathrm{bce}} = 0.3$, $\lambda_{\mathrm{cite}} = 0.25$, temperature $\tau = 1.0$, and
\begin{align}
    \mathcal{L}_{\mathrm{list}}(g) &= -\log \frac{\exp(s_p / \tau)}{\sum_{j=1}^{m} \exp(s_j / \tau)}, \\
    \mathcal{L}_{\mathrm{bce}}(g) &= \frac{1}{m} \sum_{j=1}^{m} \mathrm{BCE}\big(\sigma(s_j), y_j\big), \\
    \mathcal{L}_{\mathrm{cite}}(g) &= \frac{1}{|\mathcal{V}_g|} \sum_{j \in \mathcal{V}_g} \frac{1}{K_j} \sum_{k=1}^{K_j} \mathrm{BCE}\big(\sigma(z_{j,k}), r_{j,k}\big).
\end{align}
Here $\mathcal{V}_g$ indexes only candidates with verified fine-grained annotations (tables and document regions), masking $\mathcal{L}_{\mathrm{cite}}$ on image-level and passage-level samples so coarse supervision never corrupts pointer weights. Negative document regions confirmed to contain no supporting field are supervised with all-zero targets $\mathbf{r}_j = \mathbf{0}$.

\paragraph{Null-Anchored Calibration Extension (\texttt{SieveVL-0.8B (Null-Cal)})}
Standard listwise InfoNCE is shift-invariant: adding any query-dependent offset $b(q)$ to all $s_j$ leaves $\mathcal{L}_{\mathrm{list}}$ unchanged, weakening cross-query score calibration when a retrieved pool contains \emph{no} gold evidence. To study cross-query calibration directly, we formulate a \textbf{null-anchored listwise loss} by prepending a fixed virtual abstention anchor $s_0 = 0$ to every group, yielding extended logits $\tilde{\mathbf{s}} = (0, s_1, \dots, s_m)$:
\begin{equation}
    \mathcal{L}_{\mathrm{null}}(g) =
    \begin{cases}
        -\log \dfrac{\exp(s_p / \tau)}{1 + \sum_{j=1}^{m} \exp(s_j / \tau)}, & \text{if } \sum_{j=1}^m y_j > 0, \\[10pt]
        \log\!\Big(1 + \sum_{j=1}^{m} \exp(s_j / \tau)\Big), & \text{if } \sum_{j=1}^m y_j = 0.
    \end{cases}
\end{equation}
For positive-containing groups, $\mathcal{L}_{\mathrm{null}}$ jointly maximizes $s_p - s_j$ and pushes $s_p > 0$. For empty (all-negative) pools, its gradient concentrates automatically on the highest-scoring distractor, driving $\max_j s_j$ below the universal zero anchor $0$. In the \texttt{Null-Cal} variant, we train with $\lambda_{\mathrm{bce}} = 0.5$, $\lambda_{\mathrm{cite}} = 0.35$, and adaptive per-row positive unit weighting $w_+ = \min(3.0, K_{\mathrm{neg}}/\max(1, K_{\mathrm{pos}}))$ for dense multi-entity strips (remaining $1.0\times$ on 4-unit panels). The full six-stratum granularity-masked training pipeline and shift-invariant vs.\ null-anchored logit geometry are illustrated in Appendix~\ref{app:repro} (Figure~\ref{fig:supp_training_nullcal}).

\subsection{Hierarchical Strip-to-Region Inference for Full-Page Documents}
Scanned forms often contain 35--105 entities on a full page (e.g., $754{\times}1000$ in FUNSD), far beyond the 4--14 unit context of localized panels. Feeding all page entities at once degrades zero-shot localization sharply ($0.0323$ F1 in our preliminary check). Algorithm~\ref{alg:hier} resolves this without changing model weights: a full page is partitioned into $S{=}6$ overlapping horizontal layout strips (30\% vertical overlap), every entity intersecting a strip is indexed inside that strip, and the frozen SieveVL-0.8B executes one non-autoregressive pass per strip to score each strip via its support head while simultaneously producing sub-band pointer probabilities. Passing the winning strip (\texttt{band\_crop}) or its pointer-selected union box (\texttt{tight\_crop}) to the 27B reader scales fine-grained pruning to arbitrary full-page forms.

\begin{algorithm}[t]
\caption{Hierarchical Band-to-Region Pruning with Frozen SieveVL-0.8B}
\label{alg:hier}
\begin{algorithmic}[1]
\Require Full document page image $I$ of size $(W, H)$, OCR/layout boxes $\mathcal{B}$, query $q$, strips $S{=}6$, overlap $\rho{=}0.30$, mode $\in \{\texttt{band}, \texttt{tight}\}$, padding $\delta{=}8$\,px.
\State Partition vertical height $H$ into $S$ overlapping intervals $\{[y_1^{(s)}, y_2^{(s)}]\}_{s=1}^{S}$ with overlap $\rho$.
\For{$s = 1 \dots S$}
    \State Collect entities $\mathcal{B}_s \subseteq \mathcal{B}$ whose vertical center lies in $[y_1^{(s)}, y_2^{(s)}]$; compute bounding union $B_s$.
    \State Expand $\mathcal{B}_s$ to include all boxes intersecting $B_s$ (ensuring zero unindexed text); update $B_s$ and crop strip $I_s = \mathrm{Crop}(I, B_s)$.
    \State Run single forward pass $(s^{(s)}, \mathbf{p}^{(s)}) \gets \text{SieveVL-0.8B}(q, I_s, \mathcal{B}_s)$.
\EndFor
\State Select winning strip $s^\star = \arg\max_{s} s^{(s)}$.
\If{mode is \texttt{band}}
    \State \Return $I_{s^\star}$ \Comment{Band-level crop ($\sim$84.5\% visual token reduction)}
\Else
    \State Select $\hat{\mathcal{B}} = \{b_k \in \mathcal{B}_{s^\star} : p_k^{(s^\star)} \ge 0.5\}$ (fallback: top-2 units if empty).
    \State \Return $\mathrm{TightCrop}(I, \mathrm{Union}(\hat{\mathcal{B}}), \delta)$ \Comment{Region-level tight crop}
\EndIf
\end{algorithmic}
\end{algorithm}

\section{Experimental Setup}
\label{sec:setup}

\subsection{Training Corpus and Zero-Leakage Protocol}
The main \texttt{SieveVL-0.8B} checkpoint is trained for 1 epoch (395 optimizer steps, 2,080.6\,s on one A100 GPU) on 9,430 query groups drawn strictly from official training splits across five sources: MMQA~\cite{talmor2021multimodalqa} (2,500 \texttt{en/image} + 804 \texttt{en/table}), WebQA~\cite{chang2022webqa} (1,851 \texttt{en/image} + 548 \texttt{en/text}), XFUND \texttt{zh.train}~\cite{xu2022xfund} (1,511 \texttt{zh/document\_region} paired panels), Chinese TableQA (1,500 \texttt{zh/table}, downsampled to prevent table dominance), and T2Ranking~\cite{xie2023t2ranking} (716 \texttt{zh/text}). From the raw 30,965-candidate pool, the 8B teacher filters native distractors to at most 3 hard negatives per image/text group prior to training, yielding \textbf{23,892 effective training candidates} (in the Main \texttt{no\_kd} run, the KL distillation weight is set to zero while preserving this hard-negative filtered input). We enforce strict page-level (SHA-256), template-level, and source-document deduplication against the 583-pair internal validation split and all evaluation benchmarks (with question-text overlap explicitly audited on XFUND-zh in Section~\ref{sec:results}); every natural image has its textual title removed.

For the SieveVL-0.8B (Null-Cal) ablation (813 steps, 6,545.4\,s, max length 2,816 tokens), we initialize from the base backbone, retain the full unfiltered 9,430-group (30,965-candidate) pool, and add 2,300 all-negative empty-pool groups (800 native MMQA distractor pools, 800 CLIP cross-question pools with mean cosine 0.2542, 400 non-image pools, 300 XFUND strip empty pools) plus 1,200 positive Chinese XFUND 6-strip groups (4--24 entities/strip) built exclusively from 105 clean train-only pages disjoint from both \texttt{v2/validation} and \texttt{XFUND/zh.val}. No English document region ever enters training (exact corpus manifests, hyperparameters, and prompt templates appear in Appendix~\ref{app:repro}, Figure~\ref{fig:supp_training_nullcal} and Tables~\ref{tab:supp_internal_val}--\ref{tab:supp_hparams}).

\subsection{Peer-Reviewed Benchmarks and Frozen Evaluation Protocols}
Table~\ref{tab:benchmarks} summarizes our five evaluation benchmarks, all published at peer-reviewed venues (full provenance and published peer-reviewed reference anchors are cataloged in Appendix~\ref{app:sota_catalog} (Table~\ref{tab:supp_sota_overview})). Crucially, we test \textbf{zero-shot transfer} on MRAG-Bench, Visual-RAG, MP-DocVQA, and English FUNSD using the exact same 0.873B checkpoint without dataset-specific tuning.

\begin{table}[tbp]
\centering
\scriptsize
\setlength{\tabcolsep}{2.5pt}
\begin{tabular*}{\linewidth}{@{\extracolsep{\fill}}>{\raggedright\arraybackslash}p{0.20\linewidth}>{\raggedright\arraybackslash}p{0.16\linewidth}c>{\raggedright\arraybackslash}p{0.12\linewidth}>{\raggedright\arraybackslash}p{0.21\linewidth}>{\raggedright\arraybackslash}p{0.19\linewidth}@{}}
\toprule
\textbf{Benchmark / Venue} & \textbf{Granularity} & \textbf{Lang.} & \textbf{Transfer} & \textbf{Frozen Scope} & \textbf{Primary Metrics} \\
\midrule
MRAG-Bench~\cite{hu2025mragbench} (ICLR'25) & Image rerank + abstain & EN & Zero-shot & 1,353 Qs (CLIP Top-5 $\to$ 3) & 27B Acc, CI, Gold@3, MRR \\
\addlinespace[2pt]
Visual-RAG \S5.1~\cite{wu2026visualrag} (SIGIR'26) & Same-species rerank & EN & Zero-shot & Full 374 Qs, 98,345 imgs & H@1/3/5/10, NDCG, Rec@10 \\
\addlinespace[2pt]
MP-DocVQA~\cite{tito2023mpdocvqa} (ICDAR'23) & Multi-page selection & EN & Zero-shot & 238 \texttt{val} docs; 5,019 \texttt{test} & Page H@1/2, ANLS, Tok \\
\addlinespace[2pt]
XFUND \texttt{zh.val}~\cite{xu2022xfund} (ACL'22) & Intra-page tight-crop & ZH & Page held-out & 47 pages, 94 paired Qs & Cite F1, EM, ANLS, Tok \\
\addlinespace[2pt]
FUNSD \texttt{test}~\cite{jaume2019funsd} (ICDAR'19) & Panel \& full-page crop & EN & 0-shot (cross-ling.) & 42 pages, 84 paired Qs & Band H@1, F1, ANLS, Tok \\
\bottomrule
\end{tabular*}
\caption{Overview of the five peer-reviewed evaluation benchmarks spanning natural-image retrieval, multi-page visual document selection, and intra-page fine-grained evidence localization. All downstream generation experiments share the frozen local \texttt{qwen3.8-27b-fp8} reader.}
\label{tab:benchmarks}
\end{table}

\begin{itemize}
    \item \textbf{MRAG-Bench~\cite{hu2025mragbench} (ICLR 2025)}: All 1,353 official test multiple-choice questions using the dataset's native CLIP-retrieved 5-image candidate pool (756 queries contain $\ge 1$ gold image in the pool; 597 contain none). Every reranker selects Top-3 images without injecting ground-truth images.
    \item \textbf{Visual-RAG \S5.1~\cite{wu2026visualrag} (SIGIR 2026)}: All 374 official v2 biology attribute questions evaluated over same-species pools of 152--300 raw iNaturalist~2021 images (98,345 unique images, 98,945 candidate pairs, mean 264.6/query; 373 complete official pools plus 1 documented source exception on query \texttt{v2-19} where one negative path is absent from \texttt{iNat21 train.tar.gz}). Rerankers re-order \texttt{CLIP-ViT-L-336} Top-20 candidates under exact SIGIR~2026 Table~2 metrics (Hit@$k$, NDCG@10, Recall@10), supplemented by a 50-question (12,818-image) multi-retriever (\texttt{CLIP} and \texttt{Qwen3-VL-Embedding-8B}), \texttt{Null-Cal}, and uninterrupted same-GPU timing diagnostic.
    \item \textbf{MP-DocVQA~\cite{tito2023mpdocvqa} (ICDAR 2023 / Pattern Recognition)}: For controlled multi-selector comparisons with accessible gold page labels, we deterministically freeze 1 question per unique 4--10 page document from official \texttt{val} (238 distinct documents, 1,459 pages, mean 6.13 pages/doc; all VLM selectors receive identical 768\,px inputs and the 27B reader receives 1024\,px inputs). To verify external generalization at full scale without label access, we additionally evaluate the exact frozen zero-shot SieveVL-0.8B (Top-2) + 27B pipeline on all 5,019 official hidden \texttt{test} questions (959 documents, 25,630 question--page pairs, 1--20 pages) via the official ICDAR RRC Task~4 evaluation server (Appendix~\ref{app:mpdoc_test}).
    \item \textbf{XFUND (\texttt{zh.val})~\cite{xu2022xfund} (ACL 2022 Findings)}: 47 unseen Chinese form pages (94 paired field questions, zero page or template overlap with training).
    \item \textbf{FUNSD (\texttt{test})~\cite{jaume2019funsd} (ICDARW 2019)}: Official first-author release (\texttt{testing\_data}, excluding macOS metadata), yielding 42 eligible English form pages (84 paired questions). We evaluate both paired panels and uncropped \textbf{full pages ($754{\times}1000$)} under zero-shot cross-lingual transfer.
\end{itemize}

\subsection{Baselines, Reader Configuration, and Statistical Testing}
To ensure peer-reviewed comparability, we exclude unreviewed preprints and compare directly against: (1) \textbf{CLIP-ViT-L-336}~\cite{radford2021clip} (0.43B) and \textbf{Qwen3-VL-Embedding-8B} (8.77B); (2) \textbf{ACL 2026 Utility-VES}~\cite{luo2026utility}, reproduced verbatim from the official open-source prompts and single-step \texttt{True}/\texttt{False} logit extraction using \texttt{Qwen3-VL-2B-Instruct} (2.13B, 2.44$\times$ larger than SieveVL-0.8B), reporting both its paper-default \texttt{true\_logit} score and its softmax-normalized \texttt{p\_true} variant; (3) \textbf{Qwen3-VL-Reranker-8B} (8.77B, 10.04$\times$ larger, \texttt{yes}--\texttt{no} logit); and (4) \textbf{Query-Only}, \textbf{All-Pages}, \textbf{Gold Oracle}, and \textbf{Counterfactual Other-Field Crop} bounds. All downstream generation runs use the identical local \texttt{qwen3.8-27b-fp8} reader (deterministic decoding, \texttt{temperature}=0.0, thinking disabled), with exact API-reported \texttt{image\_tokens} recorded per request. All timing comparisons are measured on the same dedicated NVIDIA A100 MIG \texttt{3g.40gb} slice. Document and form QA answers are evaluated via strict Exact Match (EM) and Average Normalized Levenshtein Similarity (ANLS)~\cite{biten2019scene}. Statistical significance is assessed via 2,000-replicate paired cluster-bootstrap 95\% confidence intervals grouped by query image ID (MRAG-Bench), document ID (MP-DocVQA), or page ID (FUNSD).

\section{Results and Analysis}
\label{sec:results}

\subsection{Natural-Image RAG Reranking on MRAG-Bench}
\begin{figure*}[tbp]
    \centering
    \includegraphics[width=\textwidth]{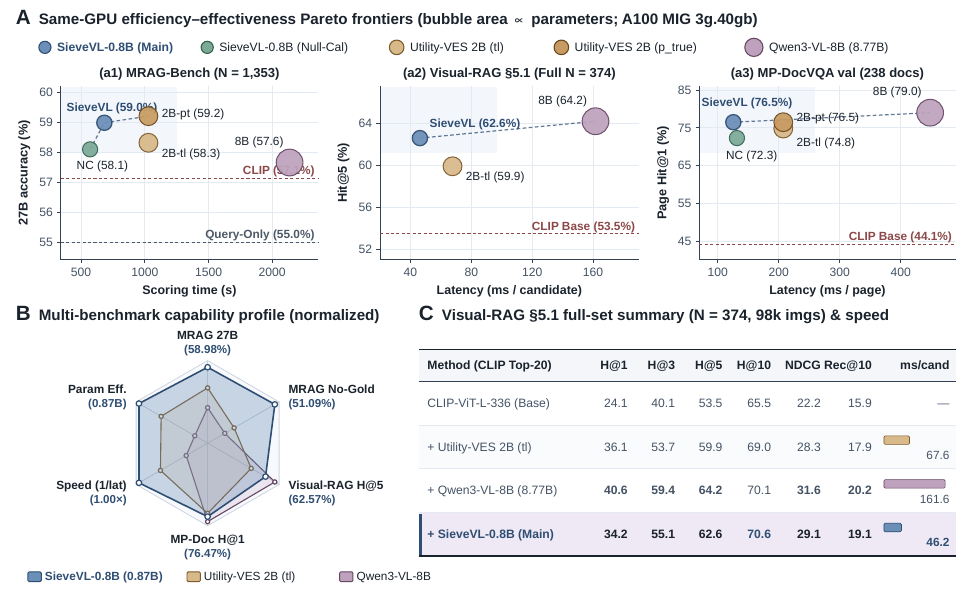}
    \caption{Cross-benchmark efficiency--effectiveness synthesis on identical hardware (NVIDIA A100 MIG \texttt{3g.40gb}). \textbf{(A)} Same-GPU Pareto frontiers with bubble area proportional to parameter count across (a1) MRAG-Bench ($N{=}1{,}353$, 27B accuracy), (a2) full Visual-RAG \S5.1 ($N{=}374$, Hit@5), and (a3) zero-shot MP-DocVQA \texttt{val} (238 docs, Page Hit@1). \textbf{(B)} Normalized six-axis multi-benchmark capability radar (scaled to $[0.18, 0.96]$ per axis relative to baseline-to-frontier bounds; SieveVL-0.8B raw values in parentheses) contrasting accuracy, no-gold robustness, speed, and parameter efficiency. \textbf{(C)} Full 374-question Visual-RAG \S5.1 retrieval summary paired with uninterrupted same-GPU latency bars: \textbf{SieveVL-0.8B} (0.87B) outperforms the 2.44$\times$ larger ACL~2026 Utility-VES (\texttt{tl}) on 5/6 metrics at 1.46$\times$--1.65$\times$ lower latency than 2B and 3.13$\times$--3.57$\times$ lower latency than 8B.}
    \label{fig:frontier}
\end{figure*}

\begin{figure*}[tbp]
    \centering
    \includegraphics[width=\textwidth]{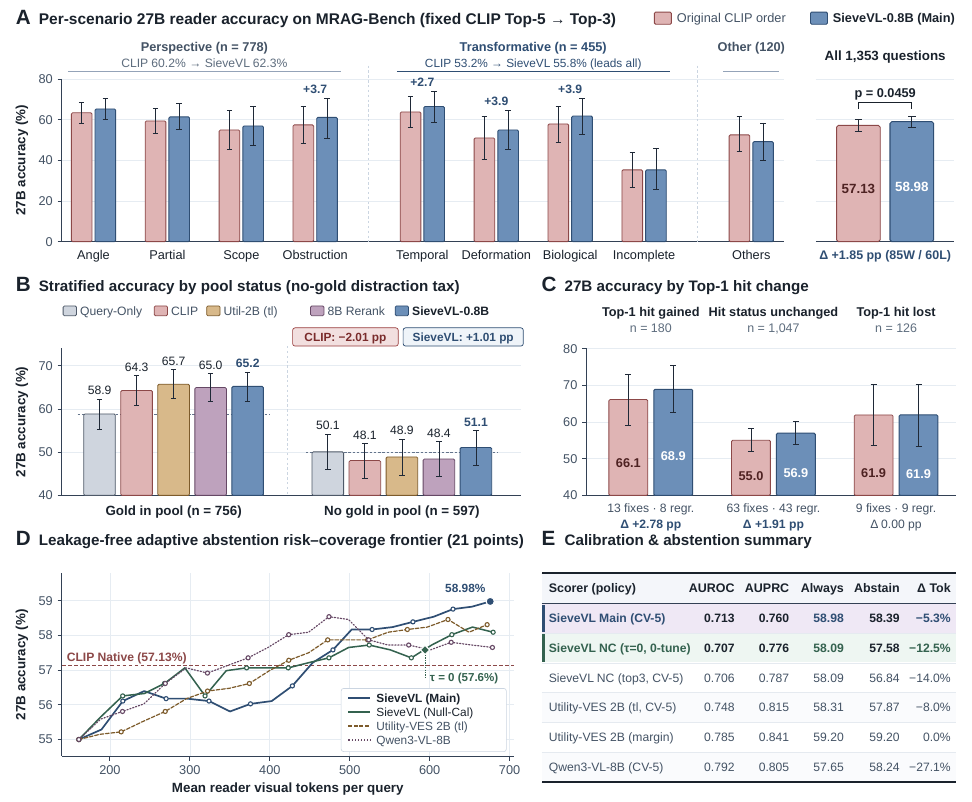}
    \caption{End-to-end MRAG-Bench panorama on the frozen 27B reader ($N{=}1{,}353$, fixed CLIP Top-5 $\to$ Top-3). \textbf{(A)} Per-scenario 27B accuracy across all nine official scenarios with 95\% paired cluster-bootstrap whiskers (grouped by query image; 4,000 resamples for per-scenario/stratified bars, 2,000 in Table~\ref{tab:mrag_main}) and two-sided exact McNemar bracket, highlighting SieveVL-0.8B's lead on Transformative queries (55.82\%). \textbf{(B)} Stratified accuracy by pool status exposing CLIP's $-2.01$\,pp no-gold distraction tax and SieveVL-0.8B's reversal of that drop (51.09\%, $+3.02$\,pp over CLIP, $+1.01$\,pp vs.\ Query-Only). \textbf{(C)} 27B accuracy grouped by Top-1 evidence-hit change (gained, unchanged, lost). \textbf{(D)} 21-point leakage-free adaptive abstention risk--coverage frontier against mean reader visual tokens. \textbf{(E)} Embedded cross-query calibration and abstention summary under query-image-disjoint 5-fold CV and zero-tuning $\tau{=}0$ (zero length truncations; complete table in Table~\ref{tab:supp_abstention}).}
    \label{fig:mrag}
\end{figure*}

Table~\ref{tab:mrag_main} reports full MRAG-Bench results with the 27B reader, while Figure~\ref{fig:frontier}(A, B) and Figure~\ref{fig:mrag}(A--C) visualize the same-GPU efficiency--effectiveness frontier, six-axis multi-benchmark capability profile, nine-scenario accuracy, and stratified no-gold performance.

\begin{table}[tbp]
\centering
\scriptsize
\setlength{\tabcolsep}{2.5pt}
\begin{tabular*}{\linewidth}{@{\extracolsep{\fill}}lrrrrrrrr@{}}
\toprule
\textbf{Method} & \textbf{Params} & \shortstack[r]{\textbf{Full}\\\textbf{(\%)}} & \shortstack[r]{\textbf{$\Delta$ vs.\ CLIP}\\\textbf{(pp, 95\% CI)}} & \shortstack[r]{\textbf{Gold}\\\textbf{(756)}} & \shortstack[r]{\textbf{No-gold}\\\textbf{(597)}} & \shortstack[r]{\textbf{Gold@3}\\\textbf{(756)}} & \textbf{MRR} & \shortstack[r]{\textbf{Score}\\\textbf{time}} \\
\midrule
Query-Only (no ret.) & -- & 54.99 & $-2.14$ ($[-4.34,+0.15]$) & 58.86 & 50.08 & -- & -- & -- \\
CLIP Native Order & 0.43B & 57.13 & $0.00$ & 64.29 & 48.07 & 88.76 & 0.4141 & -- \\
\midrule
Utility-VES (\texttt{tl})~\cite{luo2026utility} & 2.13B & 58.31 & $+1.18$ ($[-0.29,+2.88]$) & 65.74 & 48.91 & 94.58 & 0.4553 & 1028s (1.51$\times$) \\
Utility-VES (\texttt{p\_true})~\cite{luo2026utility} & 2.13B & \textbf{59.20} & $+2.07$ ($[+0.43,+3.79]$) & \textbf{66.53} & 49.92 & 94.58 & 0.4624 & 1028s (1.51$\times$) \\
Qwen3-VL-Reranker-8B & 8.77B & 57.65 & $+0.52$ ($[-1.12,+2.19]$) & 64.95 & 48.41 & \textbf{97.75} & \textbf{0.4828} & 2136s (3.13$\times$) \\
\midrule
SieveVL-0.8B (+KD) & 0.87B & 56.76 & $-0.37$ ($[-1.98,+1.27]$) & 64.15 & 47.40 & 94.05 & 0.4491 & -- \\
SieveVL-0.8B (Null-Cal)$^\dagger$ & 0.87B & 58.09 & $+0.96$ ($[-0.66,+2.60]$) & 65.48 & 48.74 & 93.92 & 0.4410 & \textbf{571s} (0.84$\times$) \\
\rowcolor{sieveRow}
\textbf{SieveVL-0.8B (Main)} & \textbf{0.87B} & 58.98 & $\mathbf{+1.85}$ ($[+0.07,+3.67]$) & 65.21 & \textbf{51.09} & 92.99 & 0.4443 & 683s (1.00$\times$) \\
\bottomrule
\end{tabular*}
\caption{End-to-end evaluation on MRAG-Bench (full $N{=}1{,}353$, fixed CLIP Top-5 $\to$ Top-3) with the frozen \texttt{qwen3.8-27b-fp8} reader scored by upstream \texttt{MRAG-Bench/eval/score.py} (zero length-truncated replies; \texttt{tl} denotes canonical \texttt{true\_logit}). Confidence intervals are 95\% paired cluster-bootstrap CIs grouped by query image; $^\dagger$one normal-stop R4 reply omitting an explicit A--D letter is counted wrong. \texttt{Gold@3} is computed over the 756 gold-in-pool queries (703/756 for Main). Wall-clock scoring times span all 6,765 pairs on the same NVIDIA A100 MIG \texttt{3g.40gb} slice.}
\label{tab:mrag_main}
\end{table}

Three findings stand out. First, \textbf{SieveVL-0.8B (Main) achieves 58.98\% 27B accuracy (798/1,353)}, delivering a statistically significant $+1.85$\,pp gain over CLIP native ordering (57.13\%, 95\% CI $[+0.07,\allowbreak +3.67]$; 85 wins vs.\ 60 losses) and $+3.99$\,pp over \texttt{Query-Only} (54.99\%, CI $[+1.65, +6.37]$). It outperforms ACL~2026 Utility-VES's canonical \texttt{true\_logit} score by $+0.67$\,pp (58.31\%, $+9$ questions; CI vs.\ CLIP crosses zero; paired CI of SieveVL vs.\ \texttt{true\_logit} $[-0.97, +2.33]$), surpasses the 10.04$\times$ larger Qwen3-VL-Reranker-8B by $+1.33$\,pp (57.65\%, $+18$ questions; paired CI $[-0.37, +2.98]$), and statistically ties the 2.44$\times$ larger softmax-normalized \texttt{p\_true} variant (59.20\%, 801/1,353; $-0.22$\,pp, 3 questions, paired CI $[-1.91, +1.45]$). Second, on identical hardware, SieveVL-0.8B scores all 6,765 pairs in 682.6\,s (1.51$\times$ faster than Utility-VES 2B at 1,028.4\,s and 3.13$\times$ faster than 8B at 2,136.2\,s). Third, the stratified breakdown in Figure~\ref{fig:mrag}(B) and Top-1 hit-change decomposition in Figure~\ref{fig:mrag}(C) explain why higher retrieval MRR alone does not guarantee higher 27B accuracy: although 8B reaches 97.75\% Gold@3, it still suffers a $-1.67$\,pp distraction drop on the 597 no-gold queries (48.41\% vs.\ 50.08\% Query-Only, comparable to CLIP's $-2.01$\,pp drop to 48.07\%), whereas \textbf{SieveVL-0.8B reverses CLIP's $-2.01$\,pp no-gold distraction drop}, reaching \textbf{51.09\%} ($+3.02$\,pp over CLIP, 95\% CI $[+0.50, +5.51]$; $+2.68$\,pp over 8B; and $+1.01$\,pp vs.\ Query-Only, CI $[-2.45, +4.62]$) while lifting gold-in-pool accuracy to 65.21\%. The nine-scenario breakdown in Figure~\ref{fig:mrag}(A) (full numerical table in Table~\ref{tab:supp_mrag_scenarios}) localizes this gain to transformative queries (455 questions), where SieveVL-0.8B leads all methods outright at \textbf{55.82\%} ($+3.73$\,pp over 8B at 52.09\%, $+3.07$\,pp over Utility-VES \texttt{true\_logit}, and $+1.53$\,pp over \texttt{p\_true}), ranking first on \texttt{Temporal} (66.44\%), \texttt{Biological} (61.76\%), and \texttt{Incomplete} (35.29\%) while beating 8B by $+7.84$\,pp on \texttt{Deformation} (54.90\% vs.\ 47.06\%).

\paragraph{Upstream Official-Script Verification}
Every arm in Table~\ref{tab:mrag_main} was scored directly using the official upstream \texttt{MRAG-Bench/eval/score.py} evaluator after ensuring all 8,093 unique reader requests completed naturally with \texttt{finish\_reason=stop} (zero length-truncated replies across all 1,353 queries; identical prompt fingerprints, selected images, ordering, and input visual tokens). Only one normal-stop response in the R4 \texttt{Null-Cal} ablation omitted an explicit A--D option letter and is counted as incorrect in its 58.09\% score (full completion verification and pairwise significance CIs appear in Table~\ref{tab:supp_mrag_official_audit}).

\subsection{Fine-Grained Same-Species Retrieval on Visual-RAG \S5.1}
Table~\ref{tab:visual_rag} and Figure~\ref{fig:frontier}(A, C) report zero-shot reranking across the \textbf{full 374-question Visual-RAG \S5.1 benchmark} (98,345 unique iNat21 images, 98,945 candidate pairs, mean 264.6 images/query; Panel~A), together with our 50-question cross-retriever transfer check on \texttt{Qwen3-VL-Embedding-8B} (Panel~B; complete 12-row 50-question CLIP/Emb-8B/\texttt{p\_true}/\texttt{Null-Cal} diagnostic and 4-split verification appear in Table~\ref{tab:supp_vrag_splits}). On the full 374-question set, our CLIP-ViT-L-336 base closely aligns with the published SIGIR~2026 Table~2 anchor (Hit@1/5/10 = 24.06/53.48/65.51\%, Recall@10 = 15.92\% vs.\ published 24.33/53.74/67.65\%, 16.70\%). Reranking CLIP Top-20 candidates with \textbf{SieveVL-0.8B (Main)} delivers large, statistically significant gains on every cutoff: Hit@1 rises from 24.06\% to \textbf{34.22\%} ($+10.16$\,pp, 95\% CI $[+5.08, +15.78]$), Hit@3 from 40.11\% to \textbf{55.08\%} ($+14.97$\,pp, $[+10.16, +20.05]$), Hit@5 from 53.48\% to \textbf{62.57\%} ($+9.09$\,pp, $[+4.01, +14.17]$), Hit@10 from 65.51\% to \textbf{70.59\%} ($+5.08$\,pp, $[+1.34, +8.56]$), NDCG@10 from 22.24\% to \textbf{29.06\%} ($+6.81$\,pp, $[+5.12, +8.79]$), and Recall@10 from 15.92\% to \textbf{19.09\%} ($+3.17$\,pp, $[+1.74, +4.67]$).

Crucially, scaling from the initial 50-question probe to all 374 official pools reveals a much stronger sub-billion result: \textbf{SieveVL-0.8B (0.87B) outperforms the 2.44$\times$ larger ACL 2026 Utility-VES \texttt{true\_logit} (2.13B) on five of six metrics}---Hit@3 (55.08\% vs.\ 53.74\%, $+1.34$\,pp), Hit@5 (62.57\% vs.\ 59.89\%, $+2.67$\,pp), Hit@10 (70.59\% vs.\ 68.98\%, $+1.60$\,pp), NDCG@10 (29.06\% vs.\ 28.27\%, $+0.79$\,pp), and a statistically significant \textbf{$+1.24$\,pp gain in Recall@10} (19.09\% vs.\ 17.85\%, 95\% CI $[+0.04, +2.46]$)---while narrowing the Hit@1 gap to $-1.87$\,pp (34.22\% vs.\ 36.10\%, CI $[-6.95, +2.94]$ crosses zero). Compared with the 10.04$\times$ larger \texttt{Qwen3-VL-Reranker-8B}, SieveVL-0.8B edges 8B on Hit@10 (70.59\% vs.\ 70.05\%) and is statistically indistinguishable on Hit@5 ($-1.60$\,pp, CI $[-5.08, +1.60]$), while running \textbf{1.46$\times$ faster than 2B} (46.23 vs.\ 67.63\,ms/cand) and \textbf{3.50$\times$ faster than 8B} (161.64\,ms/cand) in uninterrupted same-GPU MIG timing. Restricting strictly to the 373 pools without the single missing iNat21 archive path (\texttt{v2-19}) shifts every metric by $\le 0.14$\,pp (complete 4-split verification in Table~\ref{tab:supp_vrag_splits}).

\begin{table}[tbp]
\centering
\scriptsize
\setlength{\tabcolsep}{2.2pt}
\begin{tabular*}{\linewidth}{@{\extracolsep{\fill}}lrrrrrrrr@{}}
\toprule
\multicolumn{9}{@{}l}{\textbf{A. Full Visual-RAG \S5.1 ($N{=}374$, 98,345 imgs; CLIP Top-20) and same-GPU latency}}\\
\textbf{Method} & \textbf{Params} & \textbf{H@1} & \textbf{H@3} & \textbf{H@5} & \textbf{H@10} & \textbf{NDCG@10} & \textbf{Rec@10} & \textbf{ms/cand} \\
\midrule
CLIP-ViT-L-336 (Base) & 0.43B & 24.06 & 40.11 & 53.48 & 65.51 & 22.24 & 15.92 & -- \\
+ Utility-VES (\texttt{tl})~\cite{luo2026utility} & 2.13B & 36.10 & 53.74 & 59.89 & 68.98 & 28.27 & 17.85 & 67.6 (1.46$\times$) \\
+ Qwen3-VL-Reranker-8B & 8.77B & \textbf{40.64} & \textbf{59.36} & \textbf{64.17} & 70.05 & \textbf{31.57} & \textbf{20.18} & 161.6 (3.50$\times$) \\
\rowcolor{sieveRow}
+ \textbf{SieveVL-0.8B (Main)} & \textbf{0.87B} & 34.22 & 55.08 & 62.57 & \textbf{70.59} & 29.06 & 19.09 & \textbf{46.2} (1.00$\times$) \\
\midrule
\multicolumn{9}{@{}l}{\textbf{B. Cross-retriever transfer check ($N{=}50$, Qwen3-VL-Emb-8B Top-20; full 50-Q diagnostic in Table~\ref{tab:supp_vrag_splits})}}\\
Qwen3-VL-Emb-8B (Base) & 8.77B & 14.0 & 34.0 & 46.0 & 58.0 & 16.61 & -- & -- \\
+ Utility-VES (\texttt{tl})~\cite{luo2026utility} & 2.13B & \textbf{32.0} & 44.0 & 52.0 & 64.0 & 23.55 & -- & 67.6 \\
+ Qwen3-VL-Reranker-8B & 8.77B & \textbf{32.0} & \textbf{58.0} & \textbf{64.0} & \textbf{68.0} & \textbf{27.76} & -- & 161.6 \\
\rowcolor{sieveRow}
+ \textbf{SieveVL-0.8B (Main)} & \textbf{0.87B} & 24.0 & 44.0 & 58.0 & 66.0 & 23.74 & -- & \textbf{46.2} \\
\bottomrule
\end{tabular*}
\par\smallskip
\begin{minipage}{\linewidth}
\footnotesize Note: Panel~A evaluates all 374 official queries (98,945 pairs, binary nDCG normalized by all pool positives; \texttt{tl} denotes canonical \texttt{true\_logit}); same-GPU timing (ms/cand) is measured uninterrupted on the 1,875-pair MIG timing union. On the 373 complete official pools (excluding \texttt{v2-19} with 1 absent iNat21 negative), SieveVL-0.8B attains 34.32/55.23/62.47/70.51\% Hit@1/3/5/10, 29.09\% NDCG@10, and 19.13\% Recall@10 (complete 4-split, \texttt{p\_true}, and \texttt{Null-Cal} diagnostics in Table~\ref{tab:supp_vrag_splits}).
\end{minipage}
\caption{Zero-shot reranking on the Visual-RAG \S5.1 same-species retrieval benchmark: (A) full official 374-question set (98,345 unique images, mean pool 264.6) with uninterrupted same-GPU latency, and (B) 50-question cross-retriever transfer on \texttt{Qwen3-VL-Embedding-8B}.}
\label{tab:visual_rag}
\end{table}

We note two honest boundaries on Visual-RAG. First, while SieveVL-0.8B outperforms 2B across Hit@3, Hit@5, Hit@10, NDCG@10, and Recall@10 on all 374 queries and edges 8B at Hit@10, the 10.04$\times$ larger 8B reranker retains a statistically significant $+6.42$\,pp advantage at strict Hit@1 (40.64\% vs.\ 34.22\%, 95\% CI $[+1.34, +11.50]$), reflecting finer species-level visual capacity at 8B scale. Second, in a supplementary 27B open-ended LLM-as-Judge check on the 50-question diagnostic subset, all three rerankers trailed CLIP Top-3 (44\% CLIP vs.\ 40\% 8B, 34\% Utility-VES~2B, 34\% SieveVL-0.8B, and 34\% Query-Only; 95\% CIs $\pm 14$--$22$\,pp). Because $N{=}50$ open-ended biology QA lacks statistical power, we treat Visual-RAG \S5.1 strictly as a fine-grained retrieval benchmark consistent with the original SIGIR 2026 Table~2 evaluation~\cite{wu2026visualrag} (full 4-split retrieval breakdown and $N{=}50$ QA audit appear in Tables~\ref{tab:supp_vrag_splits} and~\ref{tab:supp_vrag_qa}).

\subsection{Multi-Page Document Evidence Selection and Pruning on MP-DocVQA}
Table~\ref{tab:mpdocvqa}, Figure~\ref{fig:frontier}(A), and Figure~\ref{fig:compression}(A) present zero-shot multi-page document selection across 238 unique 4--10 page documents (1,459 pages). Even though SieveVL-0.8B was never trained on MP-DocVQA or multi-page English business documents, it reaches \textbf{76.47\% Page Hit@1} (182/238) and \textbf{0.8467 MRR}, outperforming CLIP (44.12\% Hit@1) by $+32.35$\,pp and surpassing ACL 2026 Utility-VES official \texttt{true\_logit} (74.79\% Hit@1, 0.8460 MRR) while running \textbf{1.65$\times$ faster} (125.61 vs.\ 207.57\,ms/page) and \textbf{3.57$\times$ faster} than 8B (448.34\,ms/page).

\begin{table}[tbp]
\centering
\scriptsize
\setlength{\tabcolsep}{2pt}
\begin{tabular*}{\linewidth}{@{\extracolsep{\fill}}lrrrrrrrr@{}}
\toprule
\textbf{Selector / arm} & \textbf{Params} & \textbf{H@1} & \textbf{H@2} & \textbf{MRR} & \textbf{ms/p} & \textbf{ANLS} & \textbf{$\Delta$ All (95\% CI)} & \textbf{Img Tok (Red.)} \\
\midrule
All-Pages (4--10p) & -- & -- & -- & -- & -- & 0.8589 & 0.00 & 4,700 (0.0\%) \\
CLIP-ViT-L-336 (Top-1) & 0.43B & 44.12 & 63.03 & 0.6307 & 27.4 & 0.5375 & $-32.13$ ($[-38.3,-25.6]$) & 758 ($-83.9\%$) \\
\midrule
Utility-VES (\texttt{tl}, Top-1)~\cite{luo2026utility} & 2.13B & 74.79 & 86.97 & 0.8460 & 207.6 & 0.8482 & $-1.07$ ($[-5.46,+3.20]$) & 763 ($-83.8\%$) \\
Utility-VES (\texttt{tl}, Top-2)~\cite{luo2026utility} & 2.13B & 74.79 & 86.97 & 0.8460 & 207.6 & 0.8758 & $+1.69$ ($[-1.55,+4.91]$) & 1,525 ($-67.6\%$) \\
Qwen3-VL-Reranker-8B (Top-1) & 8.77B & \textbf{78.99} & \textbf{91.18} & \textbf{0.8738} & 448.3 & \textbf{0.9040} & $\mathbf{+4.51}$ ($[+0.37,+8.72]$) & 766 ($-83.7\%$) \\
\midrule
\rowcolor{sieveRow}
\textbf{SieveVL-0.8B (Main, Top-1)} & \textbf{0.87B} & 76.47 & 84.87 & 0.8467 & \textbf{125.6} & 0.8275 & $-3.14$ ($[-7.48,+1.23]$) & 761 ($-83.8\%$) \\
\rowcolor{sieveRow}
\textbf{SieveVL-0.8B (Main, Top-2)} & \textbf{0.87B} & 76.47 & 84.87 & 0.8467 & \textbf{125.6} & 0.8642 & $+0.53$ ($[-2.51,+3.47]$) & 1,528 ($-67.5\%$) \\
SieveVL-0.8B (Null-Cal, Top-2) & 0.87B & 72.27 & 85.29 & 0.8273 & 131.6 & 0.8714 & $+1.25$ ($[-1.64,+4.25]$) & 1,529 ($-67.5\%$) \\
\midrule
Gold Page (Oracle) & -- & 100.0 & 100.0 & 1.0000 & -- & 0.9490 & $+9.01$ ($[+4.99,+13.6]$) & 764 ($-83.7\%$) \\
\bottomrule
\end{tabular*}
\caption{Zero-shot multi-page document evidence selection and visual-token pruning on MP-DocVQA \texttt{val} (238 unique 4--10 page documents, 1,459 total pages). All VLM selectors receive identical 768\,px inputs on the same GPU; the \texttt{qwen3.8-27b-fp8} reader receives 1024\,px inputs.}
\label{tab:mpdocvqa}
\end{table}

Downstream 27B reading reveals a major practical payoff: sending all 4--10 pages consumes 4,700.2 visual tokens (6.02\,s/query) and yields 0.8589 ANLS, whereas selecting only \textbf{Top-2 pages with SieveVL-0.8B} cuts visual tokens by \textbf{67.5\%} (to 1,528.2 tokens, 4.92\,s) while \emph{exceeding} full-document ANLS at \textbf{0.8642} ($+0.53$\,pp; \texttt{Null-Cal} reaches \textbf{0.8714}, $+1.25$\,pp). At Top-1, SieveVL-0.8B slashes visual tokens by \textbf{83.8\%} (to 761.0) with 0.8275 ANLS (CI vs.\ All-Pages crosses zero: $[-7.48, +1.23]$), vastly outperforming CLIP Top-1 (0.5375 ANLS). Comparing against Gold Page Oracle (0.9490) and 8B Top-1 (0.9040) confirms that pruning noisy pages both reduces compute and directly relieves distraction in strong readers. Table~\ref{tab:supp_mpdoc_val_ext} decomposes why SieveVL-0.8B edges Utility-VES \texttt{true\_logit} in Page Hit@1 (76.47\% vs.\ 74.79\%) while trailing by 2.07\,pp in Top-1 ANLS: on the 27 queries hit only by 0.8B, Utility-VES still averages 0.6527 ANLS from redundant non-gold pages, whereas on the 23 queries hit only by 2B, 0.8B averages 0.3758; Top-2 reading resolves this asymmetry.

\paragraph{Official Hidden Test Leaderboard Verification (5,019 Questions)}
Submitting the exact frozen zero-shot SieveVL-0.8B (Top-2) + \texttt{qwen3.8-27b-fp8} pipeline to the official ICDAR RRC MP-DocVQA Task~4 evaluation server (all 5,019 hidden test questions, 959 documents, 25,630 question--page pairs; 66.68\% overall page pruning and 76.64\% pruning on $\ge 3$-page documents) yields \textbf{0.8755 ANLS} (numerically placing \textbf{\#5 overall} in our November 26, 2026 server snapshot) and \textbf{81.83\% (4,107/5,019) Answer-Page Prediction Accuracy}. Crucially, Answer-Page Prediction Accuracy is produced strictly by the frozen 0.873B reranker alone without invoking the 27B reader, surpassing supervised task-specific models such as Hi-VT5~\cite{tito2023mpdocvqa} (79.64\%) and peer-reviewed adaptive page-retrieval pipelines such as AVIR-Qwen2.5-VL-7B~\cite{li2025avir} (81.63\%), whereas ANLS reflects the complete decoupled 0.8B + 27B pipeline. Full leaderboard rankings, per-page-position ANLS trajectories (Pages~0--19), and comparability caveats are documented in Figure~\ref{fig:compression}(B), Appendix~\ref{app:mpdoc_test}, and Table~\ref{tab:supp_mpdoc_perpage}.

\subsection{Intra-Page Region Grounding, Full-Page Hierarchical Pruning, and Counterfactual Faithfulness}
\begin{figure*}[tbp]
    \centering
    \includegraphics[width=\textwidth]{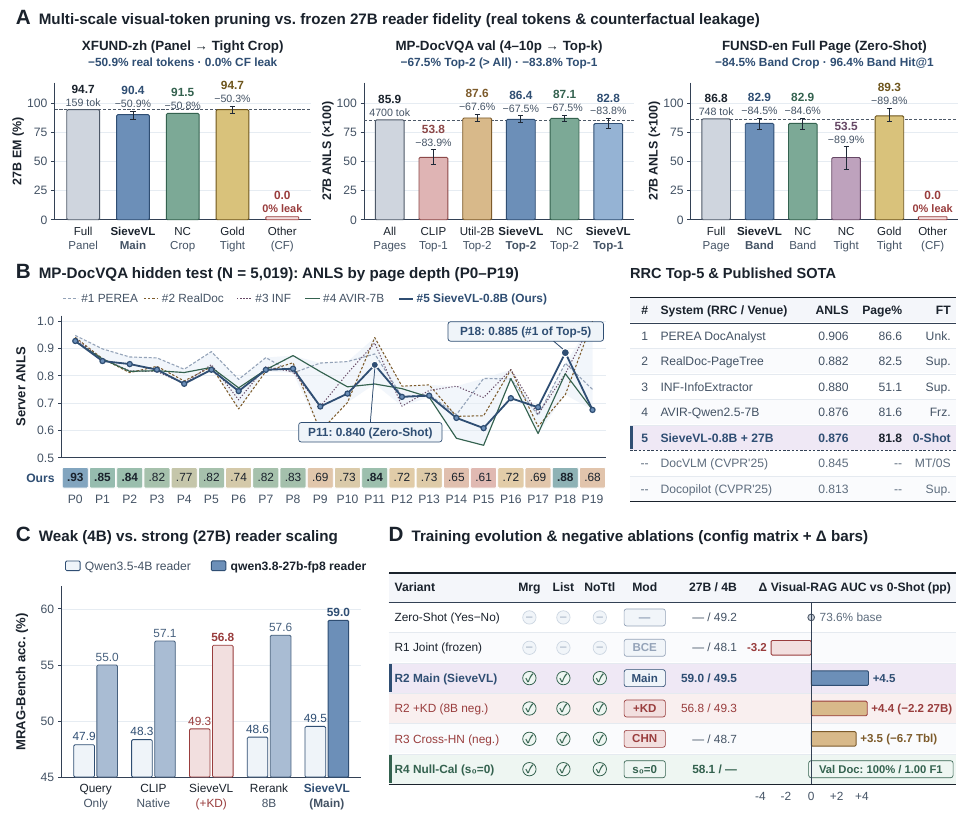}
    \caption{Document pruning, official hidden-test deep-page scaling, reader transfer, and ablation synthesis. \textbf{(A)} Multi-scale visual-token compression vs.\ frozen 27B reader fidelity (95\% paired cluster-bootstrap whiskers vs.\ uncropped input for all logged CIs; XFUND-zh \texttt{NC Crop} omits a whisker as no paired CI was recorded) across XFUND-zh ($-50.9\%$ tokens), MP-DocVQA ($-67.5\%$ at Top-2 exceeding All-Pages; $-83.8\%$ at Top-1), and zero-shot full-page English FUNSD ($-84.5\%$ real tokens, 0.0\% counterfactual leakage). \textbf{(B)} Official ICDAR RRC MP-DocVQA Task~4 hidden test ($N{=}5{,}019$, snapshot 2026-11-26) server-verified ANLS across page depths P0--P19 with heatmap strip and Top-5 / published SOTA summary (66.7\% overall page pruning; 81.83\% isolated 0.873B Top-1 Page Acc.; full matrix in Table~\ref{tab:supp_mpdoc_perpage}). \textbf{(C)} Weak (\texttt{Qwen3.5-4B}) vs.\ strong (\texttt{qwen3.8-27b-fp8}) reader scaling on MRAG-Bench. \textbf{(D)} Training evolution and retained negative ablation matrix with $\Delta$ Visual-RAG AUC bars (\texttt{Mrg}: unfrozen \texttt{visual.merger}; \texttt{List}: grouped InfoNCE; \texttt{NoTtl}: stripped image titles; \texttt{R2 +KD} drops 27B accuracy by $-2.22$\,pp; full 10-row table in Table~\ref{tab:supp_ablation}).}
    \label{fig:compression}
\end{figure*}

Unlike 2B--8B whole-image rerankers, SieveVL-0.8B localizes fine-grained visual evidence inside a page in the same forward pass. Table~\ref{tab:region_pruning} and Figure~\ref{fig:compression}(A) summarize core results on Chinese XFUND (\texttt{zh.val}) and zero-shot cross-lingual English FUNSD (\texttt{test}); full \texttt{Null-Cal} rows, scorer runtimes, 27B latencies, and 95\% confidence intervals are archived in Table~\ref{tab:supp_form_ext}.

\begin{table}[tbp]
\centering
\scriptsize
\setlength{\tabcolsep}{2pt}
\begin{tabular*}{\linewidth}{@{\extracolsep{\fill}}llrrrrrr@{}}
\toprule
\textbf{Benchmark / Arm} & \textbf{Loc.\ metric} & \shortstack[r]{\textbf{EM}\\\textbf{(\%)}} & \textbf{ANLS} & \shortstack[r]{\textbf{Abst.}\\\textbf{(\%)}} & \shortstack[r]{\textbf{Img}\\\textbf{Tok}} & \shortstack[r]{\textbf{Tok}\\\textbf{Red.}} & \shortstack[r]{\textbf{Area}\\\textbf{Red.}} \\
\midrule
\multicolumn{8}{@{}l}{\emph{A. XFUND-zh (Panel, 47 unseen pages, 94 Qs)}}\\
Full Panel & -- & 94.68 & 0.9635 & 0.00 & 158.7 & 0.0\% & 0.0\% \\
\rowcolor{sieveRow}
\textbf{SieveVL-0.8B (Pred Crop)} & \textbf{F1 0.9892} (90/94) & 90.43 & 0.9113 & 5.32 & \textbf{77.9} & \textbf{$-$50.9\%} & -- \\
Gold Tight-Crop (Oracle) & 1.0000 & 94.68 & 0.9539 & 1.06 & 78.8 & $-$50.3\% & -- \\
Other Tight-Crop (CF) & -- & 0.00 & 0.0000 & 96.81 & 78.8 & $-$50.3\% & -- \\
\midrule
\multicolumn{8}{@{}l}{\emph{B. FUNSD-en (Panel, 42 pages, 84 Qs, zero-shot cross-lingual transfer)}}\\
Full Panel & -- & 60.71 & 0.8779 & 4.76 & 76.4$^\dagger$ & 0.0\% & 0.0\% \\
\rowcolor{sieveRow}
\textbf{SieveVL-0.8B (Pred Crop)} & \textbf{F1 0.9461} (70/84) & 57.14 & 0.8398 & 8.33 & 76.5$^\dagger$ & $+$0.2\%$^\dagger$ & \textbf{$-$64.9\%} \\
Gold/Other Crop & 1.00/-- & 63.10/0.00 & 0.8932/0.0000 & 3.57/94.05 & 76.5$^\dagger$ & $+$0.1\%$^\dagger$ & $-$67.8\% \\
\midrule
\multicolumn{8}{@{}l}{\emph{C. FUNSD-en (Uncropped Full Page, $754{\times}1000$, 42 pages, 84 Qs, zero-shot)}}\\
Full Page (Uncropped) & -- & 61.90 & 0.8676 & 2.38 & 748.4 & 0.0\% & 0.0\% \\
\rowcolor{sieveRow}
\textbf{SieveVL-0.8B (Band Crop)} & \textbf{Band@1 96.43\%} & 57.14 & \textbf{0.8288} & 4.76 & \textbf{116.1} & \textbf{$-$84.5\%} & \textbf{$-$84.9\%} \\
SieveVL-0.8B (Tight Sub-Band) & F1 0.4169 & 28.57 & 0.4485 & 40.48 & 75.6 & $-$89.9\% & $-$97.6\% \\
Gold/Other Tight-Crop & 1.00/-- & 63.10/0.00 & 0.8932/0.0000 & 3.57/94.05 & 76.5 & $-$89.8\% & $-$98.7\% \\
\bottomrule
\end{tabular*}
\par\smallskip
\begin{minipage}{\linewidth}
\footnotesize $^\dagger$Pre-cropped FUNSD panels ($\sim$31k\,px$^2$) already sit at the \texttt{qwen3.8-27b-fp8} vision encoder's \texttt{min\_pixels} floor ($\approx$76 tokens); uncropped full-page evaluation (Panel~C) removes this floor and reveals $-$84.5\% real visual-token reduction. CF denotes counterfactual wrong-field cropping; \texttt{Null-Cal} variants, scorer latencies, and 95\% CIs appear in Table~\ref{tab:supp_form_ext}.
\end{minipage}
\caption{Core intra-page region grounding and evidence-cropped reading with \texttt{qwen3.8-27b-fp8} on XFUND (\texttt{zh.val}, 94 Qs, page-held-out) and FUNSD (\texttt{test}, 84 Qs, zero-shot cross-lingual transfer from Chinese-only region training; full-page evaluation in Panel~C operates on dataset OCR/layout entity boxes). Counterfactual \texttt{Other Crop} yields 0.0\% accuracy and $>$94\% abstention across every setting.}
\label{tab:region_pruning}
\end{table}

\paragraph{Held-Out Chinese Grounding (XFUND-zh)}
On 47 page- and template-disjoint unseen pages (94 paired field questions), SieveVL-0.8B attains 94/94 support accuracy, 90/94 joint exact region-set match (44/47 pages exact on both questions; 38/38 on common field question texts shared with Chinese training forms and 52/56 on strictly unseen question texts), and \textbf{0.9892 Citation F1} in 20.75\,s. Feeding only predicted tight crops to the 27B reader cuts real visual tokens by \textbf{50.9\%} ($158.7 \to 77.9$) while retaining 90.43\% Exact Match (\texttt{Null-Cal} improves 27B EM to \textbf{91.49\%}, 86/94), compared with 94.68\% for both Full Panel and Gold Tight-Crop. Counterfactual \texttt{Other Tight-Crop} drops accuracy to \textbf{0.00\%} (96.81\% abstention), proving that reader answers depend causally on the localized pixels.

\paragraph{Zero-Shot Cross-Lingual Transfer and Full-Page Pruning (FUNSD-en)}
Despite zero English region supervision during training, SieveVL-0.8B achieves 84/84 support accuracy, 70/84 joint exact match, and \textbf{0.9461 Citation F1} on English FUNSD panels (\texttt{Null-Cal} improves this to \textbf{0.9521 F1}, 71/84 exact). Because pre-cropped FUNSD panels average only 31,184\,px$^2$, cropping by 64.9\% pixel area hits the 27B encoder's minimum-token floor ($\approx 76$ tokens). Moving to \textbf{uncropped full pages ($754{\times}1000$, 748.4 visual tokens)} eliminates this floor: given page OCR/layout entity boxes, Algorithm~\ref{alg:hier} selects the correct layout band containing both key and value boxes on \textbf{96.43\% (81/84)} of questions with SieveVL-0.8B, slashing real 27B visual tokens by \textbf{84.5\%} ($748.4 \to 116.1$, reader latency $1.106 \to 0.888$\,s) while holding 27B ANLS at \textbf{0.8288} (CI vs.\ Full Page $[-8.78, +0.73]$, statistically indistinguishable from full-page reading; Gold Tight-Crop achieves 0.8932, $+2.56$\,pp over full page). Sub-band tight cropping inside 15-entity full-width English strips remains weaker (0.4485 ANLS for Main, improved by $+8.69$\,pp to 0.5354 with \texttt{Null-Cal} after adding Chinese training strips), establishing band-level pruning as the optimal zero-shot full-page operating point (complete \texttt{Null-Cal} rows, scorer runtimes, 27B latencies, and 95\% CIs appear in Table~\ref{tab:supp_form_ext}).

\subsection{Cross-Query Calibration and Leakage-Free Adaptive Abstention}
Figure~\ref{fig:mrag}(D, E) (with complete numerical metrics archived in Table~\ref{tab:supp_abstention}) evaluates whether confidence-based abstention (falling back to \texttt{Query-Only} when $\max_i s_i < \tau$) improves the accuracy--visual-token trade-off on MRAG-Bench without test-set leakage.

The comparison exposes a clear structural distinction across model scales and objectives:
\begin{itemize}
    \item \textbf{High retrieval utility raises the abstention bar.} With complete zero-truncation decoding, SieveVL-0.8B outperforms \texttt{Query-Only} by $+3.99$\,pp (58.98\% vs.\ 54.99\%) and already exceeds \texttt{Query-Only} on no-gold pools (51.09\% vs.\ 50.08\%). Consequently, falling back to \texttt{Query-Only} trades modest accuracy for visual-token savings: unnormalized ACL~2026 \texttt{true\_logit} drops $-0.44$\,pp under 5-fold CV ($58.31\% \to 57.87\%$, $-8.0\%$ tokens), while \texttt{margin} selects 0\% abstention (preserving 59.20\%).
    \item \textbf{Null-anchored calibration enables zero-tuning abstention and strong low-coverage precision.} SieveVL-0.8B (Null-Cal) abstains directly at $\tau{=}0$ without any validation tuning, cutting 12.5\% of visual tokens (16.6\% abstention) at 57.58\% accuracy ($-0.52$\,pp vs.\ always-retrieve; still $+0.44$\,pp above CLIP, CI $[-1.26, +2.19]$). At just 25\% retrieval coverage ($-56.8\%$ visual tokens, Figure~\ref{fig:mrag}(D)), \texttt{Null-Cal} already reaches 57.06\%, essentially matching full CLIP retrieval (57.13\%) at less than half the visual tokens; meanwhile, Main SieveVL-0.8B retains 58.76\% at 90\% coverage ($-6.9\%$ tokens) and 58.54\% at 85\% coverage ($-10.5\%$ tokens).
    \item \textbf{Honest 0.8B vs.\ 8B calibration gap.} Under strict 5-fold CV, 8B's higher gold-in-pool AUROC (0.7915 vs.\ 0.7127/0.7065 for 0.8B) makes it the only scorer to improve over its own always-retrieve baseline ($57.65\% \to 58.24\%$, $+0.59$\,pp, $-27.1\%$ tokens; Pareto peak 58.54\% at 60\% coverage, $-30.1\%$ tokens)---yet even 8B's best Pareto peak (58.54\%) remains below SieveVL-0.8B (Main)'s always-retrieve accuracy (58.98\%). Adding off-topic empty training pools (mean CLIP cosine 0.2542) raised 0.8B AUPRC ($0.7597 \to 0.7759$, \texttt{top3} 0.7866) without lifting AUROC toward 8B, because MRAG-Bench no-gold pools consist of fine-grained same-category distractors rather than off-topic negatives.
\end{itemize}

\subsection{Training Evolution, Retained Negative Ablations, and Reader Scaling}
Figure~\ref{fig:compression}(C, D) traces every design choice from zero-shot initialization through four training rounds across a weak (\texttt{Qwen3.5-4B}) and strong (\texttt{qwen3.8-27b-fp8}) reader, alongside the 583-pair five-stratum internal validation split (complete 10-row numerical summary in Table~\ref{tab:supp_ablation}; per-stratum internal validation in Table~\ref{tab:supp_internal_val}).

Three ablation lessons directly inform practical system design:
\begin{enumerate}
    \item \textbf{Unfreezing \texttt{visual.merger}, stripping image titles, and grouped listwise training are essential.} Moving from Round 1 (frozen visual merger, pairwise BCE, table-heavy mix) to Round 2 (\texttt{SieveVL-0.8B}) raises MRAG 4B accuracy from 48.12\% to 49.52\% and boosts Visual-RAG 12-image Pair AUC from 0.7045 to 0.7813 ($+7.68$\,pp), while reaching 100\% pair accuracy on four of five internal strata (\texttt{zh/text}: 95.56\%, Brier 0.0031--0.0440).
    \item \textbf{Distilling 8B logits (+KD) improves ranking metrics but hurts downstream 27B accuracy.} Adding KL distillation from Qwen3-VL-Reranker-8B raises MRAG Gold@3 ($92.99\% \to 94.05\%$) and MRR ($0.4443 \to 0.4491$), yet significantly drops 27B end-to-end accuracy from 58.98\% to 56.76\% ($-2.22$\,pp, 95\% CI $[-3.67, -0.81]$, falling below CLIP at 57.13\%), inheriting the 8B teacher's higher distraction on no-gold pools (47.40\% vs.\ 51.09\%).
    \item \textbf{Replacing native hard negatives with weak cross-corpus negatives (\texttt{Cross-HN}) degrades all tasks.} Substituting one native distractor with two embedding-mined cross-corpus negatives (mean cosine 0.5256 over 8,791 training images) lowers MRAG 4B accuracy ($49.52\% \to 48.71\%$), Visual-RAG Hit@3 ($66.0\% \to 60.0\%$), and internal table exactness ($82.95\% \to 76.30\%$). Native in-query distractors must be preserved intact.
\end{enumerate}

\subsection{Qualitative End-to-End Walkthrough on Real Benchmarks}
\begin{figure*}[tbp]
    \centering
    \includegraphics[width=\textwidth]{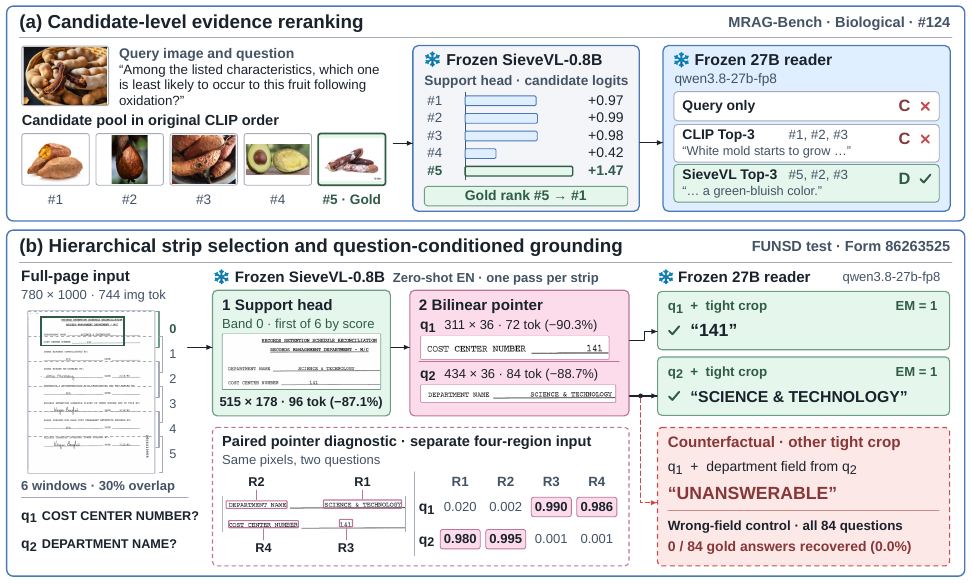}
    \caption{End-to-end real-data walkthrough through frozen \textbf{SieveVL-0.8B} and the decoupled 27B reader (\texttt{qwen3.8-27b-fp8}). \textbf{(a)} Candidate-level reranking on MRAG-Bench case~\#124 (\texttt{Biological}): SieveVL-0.8B elevates the sole gold image from CLIP rank~5 to rank~1 ($+1.47$ support logit), flipping the 27B reader from incorrect (\texttt{C} under both \texttt{Query-Only} and \texttt{CLIP Top-3}) to correct (\texttt{D}). \textbf{(b)} Zero-shot intra-page hierarchical pruning and causal verification on FUNSD test form~\texttt{86263525}: support scoring selects \texttt{Band~0} ($744 \to 96$ image tokens), pointer cropping trims to 72 and 84 tokens with exact answers (``141'' and ``SCIENCE \& TECHNOLOGY''), paired four-region diagnostics (\texttt{R1}--\texttt{R4}) shift cleanly with the query, and counterfactual cross-field cropping yields ``UNANSWERABLE''.}
    \label{fig:walkthrough}
\end{figure*}

Figure~\ref{fig:walkthrough} traces two unmodified benchmark instances end-to-end through the frozen SieveVL-0.8B checkpoint and the decoupled \texttt{qwen3.8-27b-fp8} reader, illustrating both granularities in practice. On MRAG-Bench test case~\#124 (\texttt{Biological}, Figure~\ref{fig:walkthrough}(a)), CLIP places the sole gold image at rank~5 (outside its Top-3), causing both \texttt{Query-Only} and \texttt{CLIP Top-3} to predict the wrong option (\texttt{C}); SieveVL-0.8B assigns the gold image the highest support logit ($+1.47$ vs.\ $+0.42$--$+0.99$), promoting it from rank~5 to rank~1 so the 27B reader answers \texttt{D} correctly. On uncropped zero-shot English FUNSD test form~\texttt{86263525} ($780{\times}1000$, 744 image tokens; Figure~\ref{fig:walkthrough}(b)), hierarchical band selection isolates \texttt{Band~0} (96 tokens, $-87.1\%$) and sub-band pointer crops prune inputs further to 72 and 84 tokens ($-90.3\%$ and $-88.7\%$) while yielding exact answers (\texttt{141} and \texttt{SCIENCE \& TECHNOLOGY}); paired same-page pointer diagnostics confirm sharp query-conditioned shifts across regions \texttt{R1}--\texttt{R4}, and counterfactual cross-field cropping produces \texttt{UNANSWERABLE} (0.0\% leakage).

\section{Discussion, Limitations, and Conclusion}
\label{sec:conclusion}

\subsection{Scope Boundaries and Limitations}
To maintain strict evidence boundaries, we highlight five limitations of the present study:
\begin{enumerate}
    \item \textbf{Second-stage reranker, not an ANN replacement.} SieveVL-0.8B is designed to rerank coarse candidate sets (Top-5 to Top-20 images, 4--10 document pages, or 6 layout strips). Exhaustively scanning a 1,728-image corpus in an earlier probe incurred 55.9\,s P95 latency, confirming that dense indexing remains necessary for first-stage recall.
    \item \textbf{Cross-query calibration capacity.} While \texttt{SieveVL-0.8B} excels at within-query MRAG-Bench reranking (58.98\%) and \texttt{Null-Cal} enables zero-tuning $\tau{=}0$ abstention, 0.8B gold-in-pool AUROC plateaus near 0.71 versus 0.79 for 8B. Closing this gap likely requires mining fine-grained same-category empty-pool negatives rather than generic cross-question pools.
    \item \textbf{Vision-encoder \texttt{min\_pixels} floor and dense sub-band grounding.} Visual-token savings materialize only when source images exceed the reader's minimum token floor (76 tokens on \texttt{qwen3.8-27b-fp8}), as shown by comparing FUNSD panels (31k\,px$^2$, 0\% token drop despite $-64.9\%$ area) against full FUNSD pages (764k\,px$^2$, $-84.5\%$ real tokens). Inside dense 15-entity zero-shot English strips, sub-band tight cropping still trails band cropping (0.4485 Main / 0.5354 \texttt{Null-Cal} vs.\ 0.8288 ANLS), and our full-page FUNSD evaluation relies on dataset-provided OCR/layout entity boxes with serial scorer and reader timings logged separately; integrating automated layout detectors and measuring end-to-end concurrent pipeline latency remain natural next steps.
    \item \textbf{Visual-RAG open-ended QA power.} While our Visual-RAG \S5.1 retrieval evaluation now spans all 374 official questions (98,345 unique images; 373 complete official pools plus one missing iNat21 source path on \texttt{v2-19}), full-set 374-question end-to-end open-ended QA remains future work beyond our 50-question diagnostic audit.
    \item \textbf{Single-reader focus for primary tables.} Although Figure~\ref{fig:compression}(C) (Table~\ref{tab:supp_ablation}) validates consistent gains across both 4B and 27B readers on MRAG-Bench (with all 27B responses completing cleanly at \texttt{finish\_reason=stop} and verified by upstream \texttt{score.py}), extending downstream pruning evaluations to additional closed- and open-weight MLLM families is a natural next step.
\end{enumerate}

\subsection{Conclusion}
We introduced \textbf{SieveVL-0.8B}, a unified 0.873B vision--language evidence reranker and localizer that operates in a single non-autoregressive forward pass ahead of decoupled strong multimodal readers. By combining unfrozen visual-merger adaptation, language-model LoRA, \texttt{Yes}--\texttt{No}-initialized support scoring, and a marker-indexed bilinear pointer head under a balanced six-stratum grouped objective, one frozen sub-billion checkpoint spans natural-image RAG, multi-page visual documents, and zero-shot cross-lingual scanned forms. Across five peer-reviewed benchmarks with a frozen 27B reader, SieveVL-0.8B outperforms a 2.44$\times$ larger ACL~2026 canonical utility reranker (58.31\%) and a 10.04$\times$ larger 8B reranker (57.65\%) on MRAG-Bench (58.98\%, $+1.85$\,pp over CLIP, 1.51$\times$--3.13$\times$ faster, while statistically tying the 2B \texttt{p\_true} variant at 59.20\%), surpasses the 2B baseline across all 374 Visual-RAG \S5.1 retrieval pools (62.57\% vs.\ 59.89\% Hit@5; 19.09\% vs.\ 17.85\% Recall@10) and zero-shot MP-DocVQA Page Hit@1 (76.47\%, 1.65$\times$ faster), and cuts 50.9\%--84.5\% of real 27B visual tokens across XFUND, MP-DocVQA, and full-page FUNSD with verified counterfactual faithfulness (0.0\% leakage). On the official 5,019-question MP-DocVQA hidden test server (2026-11-26 snapshot), the exact same frozen zero-shot Top-2 pipeline achieves 0.8755 ANLS (numerically placing \#5 overall) and 81.83\% isolated 0.873B page accuracy while pruning 66.68\% of pages. These results demonstrate that compact, dual-granularity evidence sieving is a practical and modular alternative to scaling monolithic rerankers or feeding unpruned visual context to strong multimodal generators.


\section*{Reproducibility and Acknowledgements}
All evaluation benchmarks (MRAG-Bench, Visual-RAG, MP-DocVQA, XFUND, and FUNSD) and base pretrained weights are publicly available from their original sources. Training manifests, evaluation scripts, and per-query audit receipts will be released upon publication. This work was supported by local high-performance GPU computing infrastructure.

\setlength{\bibsep}{2.5pt plus 0.5pt}
\bibliographystyle{unsrtnat}
\bibliography{references}

\clearpage
\appendix
\renewcommand{\thetable}{\thesection.\arabic{table}}
\renewcommand{\thefigure}{\thesection.\arabic{figure}}
\makeatletter
\@addtoreset{table}{section}
\@addtoreset{figure}{section}
\makeatother

\section{Official MP-DocVQA Hidden Test Leaderboard Evaluation (Task 4, \texorpdfstring{$N{=}5{,}019$}{N=5,019})}
\label{app:mpdoc_test}

To complement our controlled 238-document multi-selector comparison on MP-DocVQA \texttt{val} (Table~\ref{tab:mpdocvqa}), we executed the exact frozen zero-shot \textbf{SieveVL-0.8B (Main, Top-2) + \texttt{qwen3.8-27b-fp8}} pipeline on all 5,019 hidden official test questions (959 unique documents, 25,630 question--page pairs spanning 1--20 pages; mean 5.11 pages/question: 1,497 single-page, 918 two-page, and 2,604 questions on $\ge 3$-page documents averaging 8.56 pages) and uploaded the verified submission to the official ICDAR Robust Reading Competition (RRC) Task~4 evaluation server (\url{https://rrc.cvc.uab.cat/?ch=17&com=evaluation&task=4&f=1&e=1}; recorded November 26, 2026).

\paragraph{Page Compression and Strict Attribution}
Retaining at most Top-2 pages per question reduces the total pages read by the 27B model from 25,630 to 8,541 (mean 1.70 pages/question), achieving a \textbf{66.68\% overall page reduction} and a \textbf{76.64\% reduction on $\ge 3$-page documents} (22,297 $\to$ 5,208 pages). Two strict evidence boundaries govern Table~\ref{tab:app_mpdoc_leaderboard}:
\begin{enumerate}
    \item \textbf{Isolated 0.873B page-selection attribution.} The submitted answer-page index (\texttt{answer\_page}) is purely the Top-1 prediction of frozen zero-shot SieveVL-0.8B; the 27B reader is never consulted for page selection. Thus, the server-verified \textbf{81.83\% Answer-Page Prediction Accuracy (4,107/5,019)} measures zero-shot SieveVL-0.8B in isolation, ranking \textbf{\#3 among the Top-5 ANLS systems} (and \#4 across all public leaderboard entries) despite zero MP-DocVQA fine-tuning.
    \item \textbf{Heterogeneous end-to-end ANLS comparison.} Our server-verified \textbf{0.8755 ANLS} (numerically placing \textbf{\#5 overall} in the November 26, 2026 snapshot) reflects the complete decoupled 0.873B + 27B reader pipeline. Because leaderboard entries vary widely in generator size (0.3B--27B+), OCR pipelines, task-specific MP-DocVQA supervision, and peer-review status (several top entries are anonymous or unreviewed), Table~\ref{tab:app_mpdoc_leaderboard} serves as external hidden-test validation alongside peer-reviewed published anchors rather than a controlled same-reader comparison (which is preserved in Table~\ref{tab:mpdocvqa}).
\end{enumerate}

\begin{table}[tbp]
\centering
\scriptsize
\setlength{\tabcolsep}{2.5pt}
\begin{tabular*}{\linewidth}{@{\extracolsep{\fill}}rlrrll@{}}
\toprule
\textbf{Rank} & \textbf{System / method} & \textbf{ANLS} & \textbf{Page (\%)} & \textbf{Source / venue} & \textbf{MP-DocVQA FT} \\
\midrule
1 & PEREA DocAnalyst & 0.9063 & 86.63 & RRC (2026-10-09, anon.) & Unknown \\
2 & RealDoc-PageTreeIndex & 0.8823 & 82.49 & RRC (2025-11-13, unrev.) & Supervised \\
3 & INF-InfoExtractor & 0.8801 & 51.07 & RRC (2025-06-20, unrev.) & Multi-stage 7B \\
4 & AVIR-Qwen2.5-VL-7B~\cite{li2025avir} & 0.8763 & 81.63 & ACM MMAsia'25 (RRC) & Frozen LVLM$^\S$ \\
\rowcolor{sieveRow}
\textbf{5} & \textbf{SieveVL-0.8B (Top-2) + 27B} & \textbf{0.8755} & \textbf{81.83} & \textbf{Ours (RRC, 2026-11-26)} & \textbf{Zero-Shot} \\
6 & Qwen2.5-VL-7B-AWQ-lite & 0.8752 & 50.79 & RRC (2025-02-27, unrev.) & Supervised \\
\midrule
\multicolumn{6}{@{}l}{\emph{Peer-Reviewed Published SOTA Reference Systems on MP-DocVQA}}\\
\midrule
-- & DocVLM (Qwen2-VL-7B)~\cite{nacson2025docvlm} & 0.8450 & -- & CVPR 2025 (E2E) & Zero-Shot / MT \\
-- & Docopilot (InternVL2-8B)~\cite{duan2025docopilot} & 0.8130 & -- & CVPR 2025 (E2E) & Supervised \\
12 & GRAM (Large)~\cite{blau2024gram} & 0.8032 & 19.98 & CVPR 2024 (RRC) & Supervised \\
13 & GRAM C-Former~\cite{blau2024gram} & 0.7812 & 19.98 & CVPR 2024 (RRC) & Supervised \\
19 & mPLUG-DocOwl2 (8B)~\cite{hu2025docowl2} & 0.6942$^\dagger$ & 50.79 & ACL 2025 (OCR-free) & Supervised \\
-- & VDocRAG (Phi3-Vision-4B)~\cite{tanaka2025vdocrag} & 0.6260 & -- & CVPR 2025 (RAG) & Supervised \\
21 & Hi-VT5 (316M)~\cite{tito2023mpdocvqa} & 0.6184$^\ddagger$ & 79.64 & Pattern Recog.\ 2023 (RRC) & Supervised \\
22 & Longformer-base concat~\cite{tito2023mpdocvqa} & 0.5287 & 71.17 & Pattern Recog.\ 2023 (RRC) & Supervised \\
25 & LayoutLMv3-base concat~\cite{tito2023mpdocvqa,huang2022layoutlmv3} & 0.4538 & 51.94 & Pattern Recog.\ 2023 (RRC) & Supervised \\
\bottomrule
\end{tabular*}
\par\smallskip
\begin{minipage}{\linewidth}
\footnotesize Note: Submitted on RRC under \texttt{0.8B Page Reranker with a 27B Reader} (2026-11-26) using frozen \texttt{SieveVL-0.8B (Main)} (\texttt{r2\_listwise\_no\_kd}). $^\dagger$Paper reports 0.6942 (RRC lists 0.6932 as \texttt{mPLUG-DocOwls}). $^\ddagger$Paper reports 0.6201 (RRC lists 0.6184). $^\S$Li et al.~\cite{li2025avir} report 0.8458 ANLS with a 3B reader in ACM MMAsia~2025 Table~1; the RRC server entry uses the 7B configuration (\texttt{AVIR-Qwen2.5-VL-7B}, 0.8763 ANLS, 81.63\% Page Acc.).
\end{minipage}
\caption{Official ICDAR RRC MP-DocVQA Task~4 hidden test leaderboard snapshot (5,019 questions, recorded November 26, 2026) combined with peer-reviewed published SOTA systems. ``MP-DocVQA FT'' denotes whether MP-DocVQA training labels were used.}
\label{tab:app_mpdoc_leaderboard}
\end{table}

\paragraph{Deep-Page Generalization (Pages~0--19)}
Despite zero MP-DocVQA fine-tuning and 66.68\% overall page pruning, Figure~\ref{fig:compression}(B) shows that our frozen pipeline remains competitive across deep document positions (e.g., Page~2: 0.8430, Page~11: 0.8400, Page~17: 0.6850, and Page~18: 0.8846, leading all Top-5 entries on Page~18). The complete 20-position server-verified Top-5 numerical matrix is archived in Appendix~\ref{app:sota_catalog} (Table~\ref{tab:supp_mpdoc_perpage}).
\FloatBarrier

\section{Complete SOTA Benchmark Catalog, Fine-Grained Breakdowns, and Published Reference Context}
\label{app:sota_catalog}

This section consolidates full specifications, peer-reviewed reference anchors, and fine-grained error diagnostics across all five SOTA evaluation benchmarks. Table~\ref{tab:supp_sota_overview} summarizes each benchmark's official provenance, our frozen evaluation scope, and published peer-reviewed anchors with explicit protocol-comparability boundaries.

\begin{table}[tbp]
\centering
\scriptsize
\setlength{\tabcolsep}{2.5pt}
\begin{tabular*}{\linewidth}{@{\extracolsep{\fill}}>{\raggedright\arraybackslash}p{0.18\linewidth}>{\raggedright\arraybackslash}p{0.24\linewidth}>{\raggedright\arraybackslash}p{0.27\linewidth}>{\raggedright\arraybackslash}p{0.26\linewidth}@{}}
\toprule
\textbf{Benchmark / Venue} & \textbf{Official split \& scale} & \textbf{SieveVL-0.8B frozen result} & \textbf{Published anchors \& note} \\
\midrule
MRAG-Bench~\cite{hu2025mragbench}\newline (ICLR 2025) & Full \texttt{test}: 1,353 MCQs, 16,130 corpus imgs, 9 scenarios, CLIP Top-5 pool & \textbf{58.98\%} 27B Acc ($+1.85$\,pp vs.\ CLIP, 0 trunc.), 682.6\,s (1.51--3.13$\times$ faster); Transformative \textbf{55.82\%} & LLaVA-OV 50.11\%/58.98\%, GPT-4o 68.68\%/74.50\%~\cite{hu2025mragbench}; ACL~2026 Utility-VES~\cite{luo2026utility} reproduced verbatim on 27B (Table~\ref{tab:supp_mrag_official_audit}). \\
\addlinespace[4pt]
Visual-RAG \S5.1~\cite{wu2026visualrag}\newline (SIGIR 2026) & Full v2: 374 Qs, 98,345 unique imgs (98,945 pairs, 152--300/Q; 373 complete + 1 missing neg.\ \texttt{v2-19}) & Full 374 CLIP Top-20: H@1/3/5/10 = 34.22/55.08/62.57/\textbf{70.59\%}, NDCG@10 29.06\%, Rec@10 \textbf{19.09\%} (beats 2B on 5/6 metrics) & Table~2 (374 Qs)~\cite{wu2026visualrag}: CLIP 24.3/53.7/67.7\%, Rec@10 16.70\%, BGE 26.7/58.6/69.0\%, Rec@10 18.81\%. 4-split check in Table~\ref{tab:supp_vrag_splits}, $N{=}50$ QA in Table~\ref{tab:supp_vrag_qa}. \\
\addlinespace[4pt]
MP-DocVQA~\cite{tito2023mpdocvqa}\newline (PR 2023 / ICDAR) & \texttt{val}: 238 unique 4--10p docs (1,459p); \texttt{test}: 5,019 Qs, 959 docs, 25,630p (1--20p) & \texttt{val}: 76.47\% H@1, 0.8642 Top-2 ANLS ($-67.5\%$ tok); \texttt{test}: \textbf{0.8755 ANLS} (snap.\ \#5), \textbf{81.83\%} Page Acc & Hi-VT5 0.6201 (79.6\% Page), GRAM 0.8032, DocOwl2 0.6942, Docopilot 0.813, DocVLM 0.845, AVIR 0.8458 (RRC 0.8763). Same-reader \texttt{val} in Table~\ref{tab:mpdocvqa}; RRC in Appendix~\ref{app:mpdoc_test}. \\
\addlinespace[4pt]
XFUND \texttt{zh.val}~\cite{xu2022xfund}\newline (ACL 2022 Find.) & 47 unseen Chinese form pages, 94 paired key--value Qs (zero template overlap) & Cite F1 \textbf{0.9892} (90/94 exact), 27B EM/ANLS 90.43\% / 0.9113 (Null-Cal 91.49\% / 0.9220), $-50.9\%$ tok, 0.0\% counterfactual & Xu et al.~\cite{xu2022xfund} benchmarks token SER/RE (LayoutXLM); here evaluates visual crop faithfulness before a frozen 27B reader. \\
\addlinespace[4pt]
FUNSD \texttt{test}~\cite{jaume2019funsd}\newline (ICDARW 2019) & 42 eligible English form pages, 84 paired Qs (36--105 entities/page; $754{\times}1000$) & 0-shot Panel F1 0.9461; Full-Page Band@1 \textbf{96.43\%}, 0.8288 ANLS, \textbf{$-84.5\%$ tok}, 0.0\% counterfactual & Jaume et al.~\cite{jaume2019funsd} benchmarks OCR entity linking; here tests zero-shot Chinese-to-English spatial cropping transfer. \\
\bottomrule
\end{tabular*}
\caption{Consolidated SOTA benchmark catalog across all five peer-reviewed datasets, linking frozen zero-shot/held-out results to published peer-reviewed reference anchors and explicit comparability notes.}
\label{tab:supp_sota_overview}
\end{table}

\subsection{MRAG-Bench Nine-Scenario and Aspect Breakdown}
Table~\ref{tab:supp_mrag_scenarios} disaggregates the full 1,353-question MRAG-Bench 27B accuracy from Table~\ref{tab:mrag_main} across all nine official visual scenarios and three top-level aspects~\cite{hu2025mragbench}. On \textbf{Perspective} queries (778 questions: \texttt{Angle}, \texttt{Partial}, \texttt{Scope}, \texttt{Obstruction}), all learned and utility rerankers lift CLIP (60.15\%) by $+2.1$--$+3.5$\,pp (SieveVL-0.8B 62.34\%, 8B 62.21\%, Utility-VES 63.11\%--63.62\%). The decisive separation occurs on \textbf{Transformative} queries (455 questions: \texttt{Temporal}, \texttt{Deformation}, \texttt{Incomplete}, \texttt{Biological}): \textbf{SieveVL-0.8B leads every method outright at 55.82\%} (ranking first on \texttt{Temporal} at 66.44\%, \texttt{Biological} at 61.76\%, and \texttt{Incomplete} at 35.29\%, and reaching 54.90\% on \texttt{Deformation}), whereas \texttt{Qwen3-VL-Reranker-8B} drops to 52.09\% ($-3.73$\,pp behind SieveVL-0.8B) and canonical Utility-VES \texttt{true\_logit} falls to 52.75\% ($-3.07$\,pp), largely due to severe distraction on \texttt{Deformation} (47.06\% and 46.08\%, respectively).

\begin{table}[tbp]
\centering
\scriptsize
\setlength{\tabcolsep}{2.5pt}
\begin{tabular*}{\linewidth}{@{\extracolsep{\fill}}lrrrrrrrr@{}}
\toprule
\textbf{Scenario / aspect} & \textbf{$N$} & \shortstack[r]{\textbf{Query-}\\\textbf{Only}} & \textbf{CLIP} & \shortstack[r]{\textbf{Util-VES}\\\textbf{(\texttt{tl})}} & \shortstack[r]{\textbf{Util-VES}\\\textbf{(\texttt{pt})}} & \shortstack[r]{\textbf{Rerank-}\\\textbf{8B}} & \shortstack[r]{\textbf{SieveVL}\\\textbf{0.8B}} & \shortstack[r]{\textbf{Null-}\\\textbf{Cal}} \\
\midrule
Angle & 322 & 63.35 & 63.35 & 65.84 & \textbf{67.39} & 64.91 & 65.22 & 66.46 \\
Partial & 246 & 58.94 & 59.35 & 61.38 & \textbf{64.23} & 62.20 & 61.38 & 60.98 \\
Scope & 102 & 57.84 & 54.90 & \textbf{59.80} & 53.92 & 56.86 & 56.86 & 54.90 \\
Obstruction & 108 & 55.56 & 57.41 & \textbf{62.04} & 60.19 & 59.26 & 61.11 & 56.48 \\
\midrule
Temporal & 149 & 57.72 & 63.76 & 65.77 & 62.42 & 63.76 & \textbf{66.44} & \textbf{66.44} \\
Deformation & 102 & 46.08 & 50.98 & 46.08 & \textbf{56.86} & 47.06 & 54.90 & 49.02 \\
Incomplete & 102 & 29.41 & \textbf{35.29} & 33.33 & 34.31 & 33.33 & \textbf{35.29} & 34.31 \\
Biological & 102 & 57.84 & 57.84 & 59.80 & 59.80 & 58.82 & \textbf{61.76} & 60.78 \\
Others & 120 & 45.00 & \textbf{52.50} & 48.33 & 49.17 & 49.17 & 49.17 & 49.17 \\
\midrule
\emph{Aspect: Perspective} & 778 & 60.15 & 60.15 & 63.11 & \textbf{63.62} & 62.21 & 62.34 & 61.83 \\
\emph{Aspect: Transformative} & 455 & 48.79 & 53.19 & 52.75 & 54.29 & 52.09 & \textbf{55.82} & 54.07 \\
\emph{Aspect: Others} & 120 & 45.00 & \textbf{52.50} & 48.33 & 49.17 & 49.17 & 49.17 & 49.17 \\
\midrule
\textbf{Overall (Full Test)} & \textbf{1,353} & 54.99 & 57.13 & 58.31 & \textbf{59.20} & 57.65 & 58.98 & 58.09 \\
\bottomrule
\end{tabular*}
\caption{Per-scenario and per-aspect 27B reader accuracy (\%) on the full MRAG-Bench test set ($N{=}1{,}353$, fixed CLIP Top-5 $\to$ Top-3, zero length truncations). Bold marks the best retrieval-augmented arm per row.}
\label{tab:supp_mrag_scenarios}
\end{table}

\begin{table}[tbp]
\centering
\footnotesize
\setlength{\tabcolsep}{1.5pt}
\begin{tabular*}{\linewidth}{@{\extracolsep{\fill}}lrrrr@{}}
\toprule
\multicolumn{5}{@{}l}{\textbf{A. Upstream \texttt{score.py} completion audit ($N{=}1{,}353$, 0 length truncations)}}\\
\textbf{Arm} & \textbf{Correct} & \textbf{Acc.\ (\%)} & \textbf{Trunc.} & \textbf{No-opt.} \\
\midrule
Query-Only & 744 & 54.99 & 0 & 0 \\
CLIP Native Order & 773 & 57.13 & 0 & 0 \\
\midrule
Utility-VES (\texttt{tl}) & 789 & 58.31 & 0 & 0 \\
Utility-VES (\texttt{p\_true}) & \textbf{801} & \textbf{59.20} & 0 & 0 \\
Reranker-8B & 780 & 57.65 & 0 & 0 \\
\midrule
SieveVL-0.8B (+KD) & 768 & 56.76 & 0 & 0 \\
SieveVL-0.8B (Null-Cal) & 786 & 58.09 & 0 & 1 \\
\rowcolor{sieveRow}
\textbf{SieveVL-0.8B (Main)} & 798 & 58.98 & 0 & 0 \\
\bottomrule
\end{tabular*}

\medskip
\begin{tabular*}{\linewidth}{@{\extracolsep{\fill}}lrrr@{}}
\toprule
\multicolumn{4}{@{}l}{\textbf{B. Pairwise comparisons: SieveVL-0.8B (Main) vs.\ baselines}}\\
\textbf{Comparator arm} & \textbf{$\Delta$ (pp / Qs)} & \textbf{Win / Loss} & \textbf{95\% CI (pp)} \\
\midrule
vs.\ Query-Only & $+3.99$ ($+54$) & 147 / 93 & $[+1.65, +6.37]$ \\
vs.\ CLIP Native Order & $+1.85$ ($+25$) & 85 / 60 & $[+0.07, +3.67]$ \\
vs.\ Utility-VES (\texttt{tl}) & $+0.67$ ($+9$) & 65 / 56 & $[-0.97, +2.33]$ \\
vs.\ Utility-VES (\texttt{p\_true}) & $-0.22$ ($-3$) & 62 / 65 & $[-1.91, +1.45]$ \\
vs.\ Reranker-8B & $+1.33$ ($+18$) & 70 / 52 & $[-0.37, +2.98]$ \\
vs.\ SieveVL-0.8B (Null-Cal) & $+0.89$ ($+12$) & 51 / 39 & $[-0.52, +2.30]$ \\
vs.\ SieveVL-0.8B (+KD) & $+2.22$ ($+30$) & 64 / 34 & $[+0.81, +3.67]$ \\
\bottomrule
\end{tabular*}
\par\smallskip
\begin{minipage}{\linewidth}
\footnotesize Note: All 8,093 unique MRAG-Bench 27B reader requests completed with \texttt{finish\_reason=stop} (zero length-truncated replies after re-running all 502 originally length-capped payloads with identical prompt fingerprints, image order, and input token counts) and were verified directly by upstream \texttt{MRAG-Bench/eval/score.py}. One normal-stop R4 \texttt{Null-Cal} response omitted an explicit A--D letter and is counted wrong (not used by SieveVL-0.8B Main). Confidence intervals use 2,000-replicate paired cluster-bootstrap grouped by query image ID.
\end{minipage}
\caption{Complete MRAG-Bench upstream official-script verification and pairwise significance summary ($N{=}1{,}353$, 27B reader).}
\label{tab:supp_mrag_official_audit}
\end{table}

Table~\ref{tab:supp_abstention} archives the complete numerical calibration and leakage-free adaptive abstention metrics accompanying Section~\ref{sec:results} and Figure~\ref{fig:mrag}(D, E).

\begin{table}[tbp]
\centering
\scriptsize
\setlength{\tabcolsep}{2pt}
\begin{tabular*}{\linewidth}{@{\extracolsep{\fill}}lrrrrrrl@{}}
\toprule
\textbf{Scorer (signal / policy)} & \textbf{AUROC} & \textbf{AUPRC} & \textbf{Always} & \textbf{Abstain ($\Delta$)} & \textbf{Abst.} & \textbf{Tok Red.} & \textbf{Pareto highlight} \\
\midrule
\rowcolor{sieveRow}
\textbf{SieveVL-0.8B (\texttt{max}, CV-5)} & 0.7127 & 0.7597 & 58.98 & 58.39 ($-0.59$) & 7.8\% & $-5.3\%$ & 90\% cov: \textbf{58.76} ($-6.9\%$) \\
SieveVL-0.8B (NC, $\tau{=}0$ 0-tune) & 0.7065 & 0.7759 & 58.09 & 57.58 ($-0.52$) & 16.6\% & $-12.5\%$ & 25\% cov: 57.06 ($-56.8\%$) \\
SieveVL-0.8B (NC, \texttt{top3}, CV-5) & 0.7056 & 0.7866 & 58.09 & 56.84 ($-1.26$) & 18.8\% & $-14.0\%$ & 65\% cov: 57.95 ($-26.4\%$) \\
Utility-VES 2B (\texttt{tl}, CV-5) & 0.7477 & 0.8153 & 58.31 & 57.87 ($-0.44$) & 11.0\% & $-8.0\%$ & 90\% cov: 58.46 ($-7.3\%$) \\
Utility-VES 2B (\texttt{margin}, CV-5) & 0.7846 & \textbf{0.8412} & \textbf{59.20} & \textbf{59.20} ($+0.00$) & 0.0\% & $0.0\%$ & 95\% cov: 58.46 ($-3.4\%$) \\
Qwen3-VL-8B (\texttt{max}, CV-5) & \textbf{0.7915} & 0.8048 & 57.65 & 58.24 ($\mathbf{+0.59}$) & 35.8\% & $-27.1\%$ & 60\% cov: 58.54 ($-30.1\%$) \\
\bottomrule
\end{tabular*}
\caption{Complete cross-query gold-in-pool detection and leakage-free adaptive abstention (fallback to \texttt{Query-Only} below threshold) on MRAG-Bench ($N{=}1{,}353$, 27B reader; NC denotes \texttt{Null-Cal}), accompanying Section~\ref{sec:results} and Figure~\ref{fig:mrag}(D, E). Five-fold CV thresholds are selected strictly on query-image-disjoint training folds; $\tau{=}0$ requires zero validation tuning.}
\label{tab:supp_abstention}
\end{table}

\subsection{Visual-RAG \S5.1 Full-Split Retrieval Audit and Open-Ended QA Diagnostics}
Table~\ref{tab:supp_vrag_splits} verifies our full 374-question Visual-RAG \S5.1 retrieval evaluation across all four diagnostic slices---373 complete official pools (excluding query \texttt{v2-19}, whose pool lists one negative path \path{Darmera_peltata/e787b9e1-...jpg} absent from official iNaturalist~2021 \texttt{train.tar.gz} and \texttt{train.json}, whereas the identical UUID appears as a positive under \texttt{Teucrium} in the same query), all 374 available pools, the 324 newly expanded questions, and the original 50-question subset---confirming consistent sub-billion gains over CLIP and ACL~2026 Utility-VES \texttt{true\_logit} on the 324-question out-of-sample expansion as well as the full set.

\begin{table}[tbp]
\centering
\scriptsize
\setlength{\tabcolsep}{1.8pt}
\begin{tabular*}{\linewidth}{@{\extracolsep{\fill}}llrrrrrr@{}}
\toprule
\multicolumn{8}{@{}l}{\textbf{A. Four-split full-set verification (98,345 unique raw images; CLIP-ViT-L-336 Top-20)}}\\
\textbf{Split} & \textbf{Method (CLIP Top-20)} & \textbf{H@1} & \textbf{H@3} & \textbf{H@5} & \textbf{H@10} & \textbf{NDCG} & \textbf{Rec@10} \\
\midrule
\multirow{4}{*}{373 Full}
 & CLIP-L-336 (Base) & 23.86 & 39.95 & 53.35 & 65.42 & 22.21 & 15.94 \\
 & + Util-VES (\texttt{tl}) & 36.19 & 53.62 & 59.79 & 68.90 & 28.28 & 17.87 \\
 & + Reranker-8B & \textbf{40.75} & \textbf{59.25} & \textbf{64.08} & 69.97 & \textbf{31.56} & \textbf{20.20} \\
 & \cellcolor{sieveRow}+ \textbf{SieveVL-0.8B} & \cellcolor{sieveRow}34.32 & \cellcolor{sieveRow}55.23 & \cellcolor{sieveRow}62.47 & \cellcolor{sieveRow}\textbf{70.51} & \cellcolor{sieveRow}29.09 & \cellcolor{sieveRow}19.13 \\
\midrule
\multirow{4}{*}{324 New}
 & CLIP-L-336 (Base) & 24.38 & 38.89 & 52.47 & 63.58 & 22.09 & 16.95 \\
 & + Util-VES (\texttt{tl}) & 34.26 & 51.54 & 57.72 & 66.67 & 27.40 & 18.67 \\
 & + Reranker-8B & \textbf{39.81} & \textbf{58.33} & \textbf{62.04} & 68.21 & \textbf{30.90} & \textbf{21.19} \\
 & \cellcolor{sieveRow}+ \textbf{SieveVL-0.8B} & \cellcolor{sieveRow}33.64 & \cellcolor{sieveRow}53.40 & \cellcolor{sieveRow}61.11 & \cellcolor{sieveRow}\textbf{68.83} & \cellcolor{sieveRow}28.54 & \cellcolor{sieveRow}20.20 \\
\midrule
\multirow{4}{*}{50 Orig.}
 & CLIP-L-336 (Base) & 22.00 & 48.00 & 60.00 & 78.00 & 23.28 & 9.26 \\
 & + Util-VES (\texttt{tl}) & \textbf{48.00} & \textbf{68.00} & 74.00 & \textbf{84.00} & 33.89 & 12.55 \\
 & + Reranker-8B & 46.00 & 66.00 & \textbf{78.00} & 82.00 & \textbf{35.89} & \textbf{13.65} \\
 & \cellcolor{sieveRow}+ \textbf{SieveVL-0.8B} & \cellcolor{sieveRow}38.00 & \cellcolor{sieveRow}66.00 & \cellcolor{sieveRow}72.00 & \cellcolor{sieveRow}82.00 & \cellcolor{sieveRow}32.42 & \cellcolor{sieveRow}11.93 \\
\bottomrule
\end{tabular*}

\medskip
\begin{tabular*}{\linewidth}{@{\extracolsep{\fill}}lrrrrrr@{}}
\toprule
\multicolumn{7}{@{}l}{\textbf{B. 50-Q multi-retriever (CLIP \& Emb-8B), \texttt{p\_true}, \texttt{Null-Cal}, and MIG timing (12,818 imgs)}}\\
\textbf{Method} & \textbf{H@1} & \textbf{H@3} & \textbf{H@5} & \textbf{H@10} & \textbf{NDCG@10} & \textbf{ms/cand} \\
\midrule
CLIP-ViT-L-336 (Base) & 22.0 & 48.0 & 60.0 & 78.0 & 23.28 & -- \\
+ Utility-VES (\texttt{tl})~\cite{luo2026utility} & \textbf{48.0} & 66.0 & 76.0 & \textbf{84.0} & 34.14 & 67.6 (1.46$\times$) \\
+ Utility-VES (\texttt{p\_true})~\cite{luo2026utility} & \textbf{48.0} & \textbf{68.0} & \textbf{78.0} & \textbf{84.0} & \textbf{37.10} & 67.6 (1.46$\times$) \\
+ Qwen3-VL-Reranker-8B & 46.0 & 66.0 & 76.0 & \textbf{84.0} & 35.62 & 161.6 (3.50$\times$) \\
+ SieveVL-0.8B (Null-Cal) & 42.0 & 58.0 & 68.0 & 82.0 & 32.36 & 64.3 (1.39$\times$) \\
\rowcolor{sieveRow}
+ \textbf{SieveVL-0.8B (Main)} & 38.0 & 66.0 & 74.0$^*$ & 82.0 & 32.10$^*$ & \textbf{46.2} (1.00$\times$) \\
\midrule
Qwen3-VL-Emb-8B (Base) & 14.0 & 34.0 & 46.0 & 58.0 & 16.61 & -- \\
+ Utility-VES (\texttt{tl})~\cite{luo2026utility} & 32.0 & 44.0 & 52.0 & 64.0 & 23.55 & 67.6 \\
+ Utility-VES (\texttt{p\_true})~\cite{luo2026utility} & \textbf{38.0} & 56.0 & 62.0 & \textbf{68.0} & 26.91 & 67.6 \\
+ Qwen3-VL-Reranker-8B & 32.0 & \textbf{58.0} & \textbf{64.0} & \textbf{68.0} & \textbf{27.76} & 161.6 \\
+ SieveVL-0.8B (Null-Cal) & 30.0 & 48.0 & 60.0 & 66.0 & 25.27 & 64.3 \\
\rowcolor{sieveRow}
+ \textbf{SieveVL-0.8B (Main)} & 24.0 & 44.0 & 58.0 & 66.0 & 23.74 & \textbf{46.2} \\
\bottomrule
\end{tabular*}
\par\smallskip
\begin{minipage}{\linewidth}
\footnotesize Note: $^*$Panel~B scores the 1,875-pair union of CLIP and Emb-8B Top-20 in one uninterrupted MIG pass; scoring CLIP-only Top-20 in the full-set run (Panel~A, 50 Orig.) yields 72.0\% Hit@5 and 32.42\% NDCG@10 on the same 50 queries (1 near-tied image flip under \texttt{bfloat16} batch padding). On the 324-question out-of-sample expansion, SieveVL-0.8B outperforms ACL~2026 Utility-VES \texttt{true\_logit} on Hit@3/5/10, NDCG@10, and Recall@10 ($+1.53$\,pp) and edges 8B on Hit@10. Full 374-question summary appears in Table~\ref{tab:visual_rag}.
\end{minipage}
\caption{Full-split Visual-RAG \S5.1 retrieval verification (Panel~A) and complete 50-question multi-retriever (CLIP vs.\ Qwen3-VL-Embedding-8B), \texttt{p\_true}, \texttt{Null-Cal}, and same-GPU MIG timing diagnostic (Panel~B).}
\label{tab:supp_vrag_splits}
\end{table}

For completeness, Table~\ref{tab:supp_vrag_qa} reports the full supplementary 27B open-ended LLM-as-Judge QA audit on the 50 Visual-RAG \S5.1 questions alongside the published SIGIR 2026 Table~2 full-set (374-question) dense retrieval baselines (CLIP Hit@1/5/10 = 24.33/53.74/67.65\%, Recall@10 = 16.70\%, NDCG@10 = 30.98\%; BGE = 26.74/58.56/68.98\%, Recall@10 = 18.81\%, NDCG@10 = 33.40\%; VLM2Vec = 24.87/51.87/62.83\%, NDCG@10 = 30.09\%)~\cite{wu2026visualrag}. Note that our binary nDCG normalizes by the ideal ranking of all pool positives (yielding 22.24\% for CLIP vs.\ 30.98\% in the paper's multi-level/cutoff convention), whereas Hit@$k$ and Recall@$k$ are directly comparable and confirm that SieveVL-0.8B (19.09\% Recall@10, 70.59\% Hit@10) surpasses all published single-stage baselines including BGE-VL-large (18.81\%, 68.98\%). On the 50-question QA subset, every reranker yields wide confidence intervals ($\pm 14$--$22$\,pp) and trails CLIP Top-3 (44\%) in open-ended judge accuracy, confirming that statistically powered open-ended QA comparisons require full-set 374-question generation.

\begin{table}[tbp]
\centering
\footnotesize
\setlength{\tabcolsep}{2pt}
\begin{tabular*}{\linewidth}{@{\extracolsep{\fill}}>{\raggedright\arraybackslash}p{\dimexpr0.38\linewidth-4pt\relax}>{\raggedleft\arraybackslash}p{\dimexpr0.14\linewidth-4pt\relax}>{\raggedleft\arraybackslash}p{\dimexpr0.16\linewidth-4pt\relax}>{\raggedleft\arraybackslash}p{\dimexpr0.32\linewidth-4pt\relax}@{}}
\toprule
\multicolumn{4}{@{}l}{\textbf{A. Judge accuracy (Visual-RAG \S5.1, $N{=}50$)}}\\
\textbf{Reader arm} & \textbf{Judge correct} & \textbf{Judge acc. (\%)} & \textbf{$\Delta$ vs. CLIP (95\% CI)} \\
\midrule
Query-Only (no image) & 17/50 & 34.0 & $-10.0$ ($[-26.0, +4.0]$) \\
CLIP-ViT-L-336 Top-3 & 22/50 & 44.0 & $0.0$ \\
CLIP + Utility-VES 2B (\texttt{true\_logit}) & 17/50 & 34.0 & $-10.0$ ($[-20.0, 0.0]$) \\
CLIP + SieveVL-0.8B (Main) & 17/50 & 34.0 & $-10.0$ ($[-22.0, +2.0]$) \\
CLIP + Qwen3-VL-Reranker-8B & 20/50 & 40.0 & $-4.0$ ($[-14.0, +6.0]$) \\
\bottomrule
\end{tabular*}

\medskip
\begin{tabular*}{\linewidth}{@{\extracolsep{\fill}}>{\raggedright\arraybackslash}p{\dimexpr0.46\linewidth-4pt\relax}>{\raggedleft\arraybackslash}p{\dimexpr0.16\linewidth-4pt\relax}>{\raggedleft\arraybackslash}p{\dimexpr0.18\linewidth-4pt\relax}>{\raggedleft\arraybackslash}p{\dimexpr0.2\linewidth-4pt\relax}@{}}
\toprule
\multicolumn{4}{@{}l}{\textbf{B. Answer overlap and reader cost}}\\
\textbf{Reader arm} & \textbf{Token F1} & \textbf{Mean img tok} & \textbf{27B latency (s)} \\
\midrule
Query-Only (no image) & 0.1320 & 0.0 & 1.449 \\
CLIP-ViT-L-336 Top-3 & 0.0718 & 558.0 & 2.217 \\
CLIP + Utility-VES 2B (\texttt{true\_logit}) & 0.0681 & 541.1 & 2.289 \\
CLIP + SieveVL-0.8B (Main) & 0.0666 & 560.8 & 2.434 \\
CLIP + Qwen3-VL-Reranker-8B & 0.0673 & 566.1 & 2.324 \\
\bottomrule
\end{tabular*}
\caption{Supplementary open-ended 27B LLM-as-Judge QA audit on Visual-RAG \S5.1 ($N{=}50$, CLIP Top-3 input), transparently documenting the underpowered negative QA outcome noted in Section~\ref{sec:results}.}
\label{tab:supp_vrag_qa}
\end{table}

\subsection{MP-DocVQA \texttt{val} Extended Diagnostics and Form Latency Summary}
Table~\ref{tab:supp_mpdoc_val_ext} provides the remaining MP-DocVQA \texttt{val} metrics omitted from Table~\ref{tab:mpdocvqa} for space (Hit@3, strict Exact Match, and 27B wall-clock latency), together with the paired Top-1 Hit@1 vs.\ ANLS decomposition between SieveVL-0.8B and Utility-VES (\texttt{true\_logit}). Table~\ref{tab:supp_mpdoc_perpage} archives the complete server-verified ANLS matrix by ground-truth page depth (Pages~0--19) for the Top-5 ICDAR RRC MP-DocVQA Task~4 hidden-test entries accompanying Figure~\ref{fig:compression}(B) and Appendix~\ref{app:mpdoc_test}. Table~\ref{tab:supp_form_ext} archives all \texttt{Null-Cal} variants, localization metrics, abstention rates, area reductions, scorer runtimes, 27B latencies, and 95\% confidence intervals across XFUND-zh and FUNSD-en (panel and full-page).

\begin{table}[tbp]
\centering
\footnotesize
\setlength{\tabcolsep}{2pt}
\begin{tabular*}{\linewidth}{@{\extracolsep{\fill}}>{\raggedright\arraybackslash}p{\dimexpr0.36\linewidth-4pt\relax}>{\raggedleft\arraybackslash}p{\dimexpr0.11\linewidth-4pt\relax}>{\raggedleft\arraybackslash}p{\dimexpr0.14\linewidth-4pt\relax}>{\raggedleft\arraybackslash}p{\dimexpr0.14\linewidth-4pt\relax}>{\raggedleft\arraybackslash}p{\dimexpr0.14\linewidth-4pt\relax}>{\raggedleft\arraybackslash}p{\dimexpr0.11\linewidth-4pt\relax}@{}}
\toprule
\multicolumn{6}{@{}l}{\textbf{A. MP-DocVQA validation}}\\
\textbf{Reader arm} & \textbf{Hit@3} & \textbf{27B EM} & \textbf{27B ANLS} & \textbf{Img Tok} & \textbf{Lat. (s)} \\
\midrule
All-Pages (4--10p) & -- & 77.73\% & 0.8589 & 4,700.2 & 6.023 \\
CLIP Top-1 & 76.05 & 47.90\% & 0.5375 & 758.4 & 4.902 \\
Utility-VES (\texttt{tl}, Top-1) & 94.12 & 76.05\% & 0.8482 & 763.4 & 4.769 \\
Utility-VES (\texttt{tl}, Top-2) & 94.12 & 79.41\% & 0.8758 & 1,524.9 & 5.057 \\
Qwen3-VL-Reranker-8B (Top-1) & 94.96 & 80.67\% & 0.9040 & 766.0 & 4.582 \\
\textbf{SieveVL-0.8B (Main, Top-1)} & 91.60 & 75.21\% & 0.8275 & 761.0 & 4.724 \\
\textbf{SieveVL-0.8B (Main, Top-2)} & 91.60 & 77.31\% & 0.8642 & 1,528.2 & 4.922 \\
SieveVL-0.8B (Null-Cal, Top-2) & 92.02 & 78.57\% & 0.8714 & 1,529.0 & 4.833 \\
Gold Page (Oracle) & 100.0 & 86.55\% & 0.9490 & 763.9 & 4.647 \\
\bottomrule
\end{tabular*}

\medskip
\begin{tabular*}{\linewidth}{@{\extracolsep{\fill}}>{\raggedright\arraybackslash}p{\dimexpr0.44\linewidth-4pt\relax}>{\raggedleft\arraybackslash}p{\dimexpr0.14\linewidth-4pt\relax}>{\raggedleft\arraybackslash}p{\dimexpr0.21\linewidth-4pt\relax}>{\raggedleft\arraybackslash}p{\dimexpr0.21\linewidth-4pt\relax}@{}}
\toprule
\multicolumn{4}{@{}l}{\textbf{B. Paired Top-1 error decomposition}}\\
\textbf{Top-1 paired subset} & \textbf{$N$} & \textbf{0.8B ANLS} & \textbf{2B ANLS} \\
\midrule
Both Hit@1 & 155 & 0.9669 & 0.9669 \\
Only 0.8B Hit@1 & 27 & 0.8790 & 0.6527 \\
Only 2B Hit@1 & 23 & 0.3758 & 0.8605 \\
Neither Hit@1 & 33 & 0.4453 & 0.4422 \\
\textbf{All 238 Docs} & \textbf{238} & \textbf{0.8275} & \textbf{0.8482} \\
\bottomrule
\end{tabular*}
\par\smallskip
\begin{minipage}{\linewidth}
\footnotesize Note: First-page heuristic Hit@1 = 28.57\% (68/238). Utility-VES (\texttt{p\_true}) Hit@1/2/3 = 76.47/87.39/92.44\%, MRR = 0.8524; paired Top-1 CI (0.8B vs.\ 2B \texttt{tl}): $-2.07$\,pp ($[-6.51, +2.29]$, crosses zero).
\end{minipage}
\caption{Extended MP-DocVQA \texttt{val} metrics (238 unique 4--10 page docs) and paired Top-1 error decomposition between SieveVL-0.8B (76.47\% Hit@1, 0.8275 ANLS) and Utility-VES \texttt{true\_logit} (74.79\% Hit@1, 0.8482 ANLS).}
\label{tab:supp_mpdoc_val_ext}
\end{table}

\begin{table}[tbp]
\centering
\scriptsize
\setlength{\tabcolsep}{2.5pt}
\begin{tabular*}{\linewidth}{@{\extracolsep{\fill}}lrrrrr@{\hspace{14pt}}lrrrrr@{}}
\toprule
\textbf{Page} & \textbf{\#1} & \textbf{\#2} & \textbf{\#3} & \textbf{\#4} & \textbf{\#5} & \textbf{Page} & \textbf{\#1} & \textbf{\#2} & \textbf{\#3} & \textbf{\#4} & \textbf{\#5} \\
\midrule
P0 & 0.9477 & 0.9407 & 0.9342 & 0.9259 & \textbf{0.9273} & P10 & 0.8520 & 0.7023 & 0.8089 & 0.7596 & \textbf{0.7348} \\
P1 & 0.8984 & 0.8627 & 0.8619 & 0.8618 & \textbf{0.8540} & P11 & 0.8800 & 0.9400 & 0.9200 & 0.7700 & \textbf{0.8400} \\
P2 & 0.8688 & 0.8115 & 0.8178 & 0.8162 & \textbf{0.8430} & P12 & 0.7243 & 0.7621 & 0.6886 & 0.7529 & \textbf{0.7229} \\
P3 & 0.8656 & 0.8346 & 0.8158 & 0.8186 & \textbf{0.8225} & P13 & 0.7310 & 0.7667 & 0.7464 & 0.7219 & \textbf{0.7274} \\
P4 & 0.8239 & 0.7811 & 0.7716 & 0.8120 & \textbf{0.7706} & P14 & 0.6578 & 0.6515 & 0.7619 & 0.5714 & \textbf{0.6457} \\
P5 & 0.8885 & 0.8326 & 0.8408 & 0.8306 & \textbf{0.8223} & P15 & 0.7902 & 0.6538 & 0.7203 & 0.5455 & \textbf{0.6084} \\
P6 & 0.7869 & 0.6787 & 0.7120 & 0.7558 & \textbf{0.7440} & P16 & 0.7883 & 0.8225 & 0.8225 & 0.7899 & \textbf{0.7181} \\
P7 & 0.8667 & 0.8163 & 0.8329 & 0.8225 & \textbf{0.8218} & P17 & 0.6550 & 0.6144 & 0.6600 & 0.5884 & \textbf{0.6850} \\
P8 & 0.8105 & 0.8477 & 0.8147 & 0.8740 & \textbf{0.8259} & P18 & 0.8462 & 0.7308 & 0.8077 & 0.8077 & \textbf{0.8846} \\
P9 & 0.8466 & 0.5966 & 0.6875 & 0.8152 & \textbf{0.6875} & P19 & 0.7500 & 1.0000 & 1.0000 & 0.6750 & \textbf{0.6750} \\
\bottomrule
\end{tabular*}
\caption{Server-verified ANLS by ground-truth answer page position (Pages~0--19) for the Top-5 ICDAR RRC MP-DocVQA Task~4 hidden-test entries (2026-11-26), accompanying Figure~\ref{fig:compression}(B) and Appendix~\ref{app:mpdoc_test}: \#1 PEREA DocAnalyst, \#2 RealDoc-PageTreeIndex, \#3 INF-InfoExtractor, \#4 AVIR-Qwen2.5-VL-7B, and \textbf{\#5 SieveVL-0.8B (Top-2) + 27B (Ours)}. Despite zero MP-DocVQA fine-tuning and 66.68\% page pruning, our frozen pipeline remains competitive on deep positions (e.g., Page~2: 0.8430, Page~11: 0.8400, Page~18: 0.8846).}
\label{tab:supp_mpdoc_perpage}
\end{table}

\begin{table}[tbp]
\centering
\scriptsize
\setlength{\tabcolsep}{1.0pt}
\begin{tabular*}{\linewidth}{@{\extracolsep{\fill}}llrrrrrr@{}}
\toprule
\textbf{Benchmark / Arm} & \textbf{Loc.} & \shortstack[r]{\textbf{EM/ANLS}} & \shortstack[r]{\textbf{Abst.}\\\textbf{(\%)}} & \shortstack[r]{\textbf{$\Delta$ ANLS}\\\textbf{(95\% CI)}} & \shortstack[r]{\textbf{Scorer}\\\textbf{(s / ms)}} & \shortstack[r]{\textbf{27B}\\\textbf{(s)}} & \shortstack[r]{\textbf{Img tok}\\\textbf{(Red.)}} \\
\midrule
\multicolumn{8}{@{}l}{\emph{XFUND-zh Panel (47 unseen pages, 94 paired Qs)}}\\
Full Panel & -- & 94.7/.964 & 0.0 & 0.00 & -- & .596 & 158.7 (0\%) \\
\rowcolor{sieveRow}
\textbf{SieveVL (Pred)} & \textbf{.9892} & 90.4/.911 & 5.3 & $-5.2$ ($[-9.6,-1.1]$) & 20.8/221 & .506 & 77.9 ($-50.9\%$) \\
Null-Cal (Pred) & .9866 & \textbf{91.5}/.922 & 4.3 & $-4.2$ (--) & 14.4/153 & .589 & 78.1 ($-50.8\%$) \\
Gold Tight-Crop & 1.000 & 94.7/.954 & 1.1 & $-1.0$ ($[-4.2,+2.2]$) & -- & .518 & 78.8 ($-50.3\%$) \\
Other Crop (CF) & -- & 0.0/.000 & 96.8 & $-96.4$ ($[-99.8,-91.1]$) & -- & .457 & 78.8 ($-50.3\%$) \\
\midrule
\multicolumn{8}{@{}l}{\emph{FUNSD-en Panel (42 pages, 84 Qs, zero-shot; Area Red.\ $-64.9\%$/$-67.8\%$)}}\\
Full Panel & -- & 60.7/.878 & 4.8 & 0.00 & -- & .615 & 76.4 (0\%) \\
\rowcolor{sieveRow}
\textbf{SieveVL (Pred)} & .9461 & 57.1/.840 & 8.3 & $-3.8$ ($[-8.8,+0.9]$) & 12.2/146 & .544 & 76.5 ($+0.2\%$) \\
Null-Cal (Pred) & \textbf{.9521} & 58.3/.840 & 8.3 & $-3.8$ (--) & 12.1/144 & .671 & 76.1 ($-0.4\%$) \\
Gold/Other Crop & 1.0/-- & 63.1/.00 & 3.6/94 & $+1.5$/$-87.8$ & -- & .56/.50 & 76.5/76.5 \\
\midrule
\multicolumn{8}{@{}l}{\emph{FUNSD-en Full Page ($754{\times}1000$, 42 pages, 84 Qs, 0-shot; Area $-84.9$/$-85.0$/$-97.7\%$)}}\\
Full Page & -- & 61.9/.868 & 2.4 & 0.00 & -- & 1.11 & 748.4 (0\%) \\
\rowcolor{sieveRow}
\textbf{SieveVL (Band)} & \textbf{96.4\%} & 57.1/\textbf{.829} & 4.8 & $-3.9$ ($[-8.8,+0.7]$) & 74.5/887 & .888 & 116.1 ($-84.5\%$) \\
Null-Cal (Band) & 94.1\% & 57.1/\textbf{.829} & 4.8 & $-3.9$ ($[-8.8,+0.7]$) & 70.8/843 & .903 & 115.3 ($-84.6\%$) \\
SieveVL (Tight) & .4169 & 28.6/.449 & 40.5 & $-41.9$ ($[-52.0,-31.7]$) & shared & .693 & 75.6 ($-89.9\%$) \\
Null-Cal (Tight) & .4674 & 34.5/.535 & 34.5 & $-33.2$ ($[-43.7,-23.9]$) & shared & .722 & 75.6 ($-89.9\%$) \\
Gold/Other Tight & 1.0/-- & 63.1/.00 & 3.6/94 & $+2.6$/$-86.8$ & -- & .78/.71 & 76.5 ($-89.8\%$) \\
\bottomrule
\end{tabular*}
\caption{Complete intra-page grounding, \texttt{Null-Cal} variants, abstention rates, scorer runtimes, 27B reader latencies (s), and 95\% cluster-bootstrap confidence intervals (vs.\ uncropped input) on XFUND (\texttt{zh.val}, 94 Qs) and FUNSD (\texttt{test}, 84 Qs, zero-shot cross-lingual), complementing Table~\ref{tab:region_pruning}.}
\label{tab:supp_form_ext}
\end{table}

\FloatBarrier

\section{Internal Validation, Training Manifest, Hyperparameters, and Prompts}
\label{app:repro}

\begin{figure*}[tbp]
    \centering
    \includegraphics[width=\textwidth]{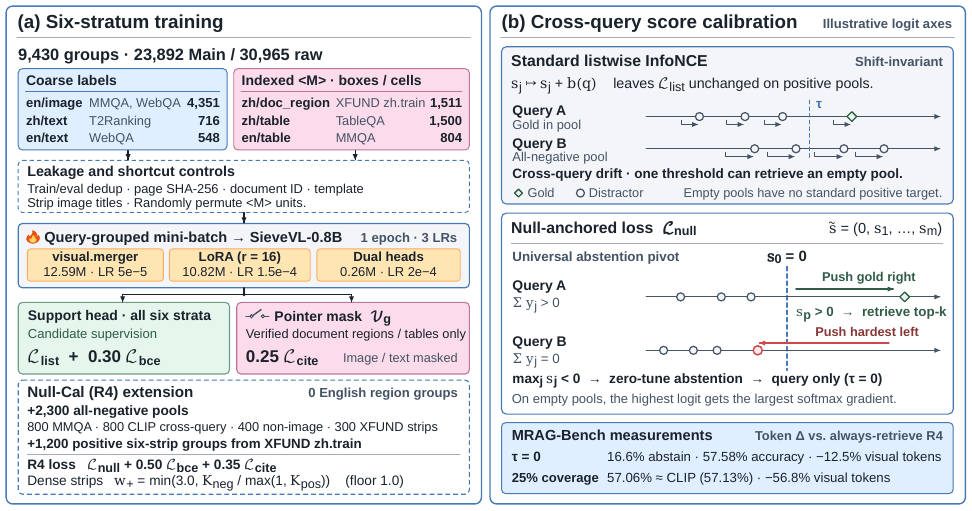}
    \caption{Six-stratum granularity-masked training and null-anchored calibration mechanism accompanying Sections~\ref{sec:method} and~\ref{sec:results}. \textbf{(a)} Six training strata (9,430 groups, 30,965 raw candidates filtered to 23,892 effective candidates for R2 Main via 8B native hard-negative selection; plus \texttt{Null-Cal} R4's full 30,965 pool, 2,300 all-negative empty pools, and 1,200 Chinese 6-strip groups, zero English regions) pass through zero-leakage deduplication, image-title stripping, and random \texttt{<M>} unit permutation into three-group differential optimization; granularity masking restricts $\mathcal{L}_{\mathrm{cite}}$ to fine-grained region/table strata $\mathcal{V}_g$ so coarse supervision never corrupts pointer weights. \textbf{(b)} Standard listwise InfoNCE is invariant to query-dependent offsets $+b(q)$, inducing cross-query drift, whereas prepending the fixed virtual abstention anchor $s_0{=}0$ in $\mathcal{L}_{\mathrm{null}}$ aligns positive pools above zero and drives the highest-scoring empty-pool distractor below zero, enabling zero-tuning abstention on MRAG-Bench ($\tau{=}0$: 16.6\% abstention, $-12.5\%$ visual tokens, 57.58\% accuracy; 25\% coverage: 57.06\% vs.\ CLIP 57.13\% at $-56.8\%$ tokens).}
    \label{fig:supp_training_nullcal}
\end{figure*}

Figure~\ref{fig:supp_training_nullcal} illustrates the six-stratum granularity-masked training pipeline, selective three-group optimization, and null-anchored calibration geometry (\texttt{Null-Cal}, $s_0{=}0$) accompanying Section~\ref{sec:method}. Table~\ref{tab:supp_ablation} archives the complete 10-row training evolution, retained negative ablation, weak-vs-strong reader scaling, and Visual-RAG 12-image Pair AUC summary visualized in Figure~\ref{fig:compression}(C, D). Table~\ref{tab:supp_internal_val} reports complete performance across all five strata of the 583-pair internal validation split (\texttt{v2/validation.jsonl}, strictly group-disjoint from all training splits). Table~\ref{tab:supp_hparams} details exact corpus composition and optimization settings for \textbf{SieveVL-0.8B (Main)} and \textbf{SieveVL-0.8B (Null-Cal)}.

\begin{table}[tbp]
\centering
\scriptsize
\setlength{\tabcolsep}{2.5pt}
\begin{tabular*}{\linewidth}{@{\extracolsep{\fill}}lrrrrrrr@{}}
\toprule
\textbf{Model / variant} & \shortstack[r]{\textbf{MRAG}\\\textbf{4B}} & \shortstack[r]{\textbf{MRAG}\\\textbf{27B}} & \shortstack[r]{\textbf{Gold@3}\\\textbf{(\%)}} & \textbf{MRR} & \shortstack[r]{\textbf{VRAG-12}\\\textbf{AUC}} & \shortstack[r]{\textbf{Val Doc}\\\textbf{(Ex.\ / F1)}} & \shortstack[r]{\textbf{Val Table}\\\textbf{Exact}} \\
\midrule
Query-Only & 47.89 & 54.99 & -- & -- & -- & -- & -- \\
CLIP Native Order & 48.34 & 57.13 & 88.76 & 0.4141 & -- & -- & -- \\
Qwen3.5-0.8B (Zero-Shot) & 49.22 & -- & 91.93 & 0.4382 & 0.7361 & (Pair 91.4\%) & (95.5\%) \\
SieveVL-0.8B (R1 Score-Only) & 49.22 & -- & 92.99 & -- & -- & -- & -- \\
SieveVL-0.8B (R1 Joint, frozen) & 48.12 & -- & 93.65 & -- & 0.7045 & -- & -- \\
\midrule
\rowcolor{sieveRow}
\textbf{SieveVL-0.8B (R2 Main, no KD)} & \textbf{49.52} & \textbf{58.98} & 92.99 & 0.4443 & \textbf{0.7813} & 98.1\% / 0.993 & 82.95\% \\
SieveVL-0.8B (+KD, neg.\ e2e) & 49.30 & 56.76 & 94.05 & 0.4491 & 0.7803 & 95.2\% / 0.986 & \textbf{83.24\%} \\
SieveVL-0.8B (R3 Cross-HN, neg.) & 48.71 & -- & 92.72 & 0.4325 & 0.7715 & 96.2\% / 0.986 & 76.30\% \\
SieveVL-0.8B (R4 Null-Cal) & -- & 58.09 & 93.92 & 0.4410 & -- & \textbf{100\% / 1.000} & 80.35\% \\
\midrule
Qwen3-VL-Reranker-8B (10.0$\times$) & 48.56 & 57.65 & \textbf{97.75} & \textbf{0.4828} & \textbf{0.8077} & -- & -- \\
\bottomrule
\end{tabular*}
\caption{Complete training evolution, retained negative ablations, weak-vs-strong reader scaling on MRAG-Bench, Visual-RAG 12-image same-species Pair AUC, and internal 583-pair validation (\texttt{zh/doc\_reg} and \texttt{zh/table}), accompanying Section~\ref{sec:results} and Figure~\ref{fig:compression}(C, D).}
\label{tab:supp_ablation}
\end{table}

\begin{table}[tbp]
\centering
\footnotesize
\setlength{\tabcolsep}{2pt}
\begin{tabular*}{\linewidth}{@{\extracolsep{\fill}}>{\raggedright\arraybackslash}p{\dimexpr0.62\linewidth-4pt\relax}>{\raggedleft\arraybackslash}p{\dimexpr0.38\linewidth-4pt\relax}@{}}
\toprule
\multicolumn{2}{@{}l}{\textbf{Zero-shot baseline}}\\
\textbf{Stratum (pairs)} & \textbf{Zero-shot pair acc.} \\
\midrule
\texttt{zh/document\_region} (104) & 91.35\% \\
\texttt{zh/table} (346) & 95.52\% \\
\texttt{en/table} (52) & 88.46\% \\
\texttt{zh/text} (45) & 73.33\% \\
\texttt{en/text} (36) & 94.44\% \\
\bottomrule
\end{tabular*}

\medskip
\begin{tabular*}{\linewidth}{@{\extracolsep{\fill}}>{\raggedright\arraybackslash}p{\dimexpr0.41\linewidth-4pt\relax}>{\raggedleft\arraybackslash}p{\dimexpr0.59\linewidth-4pt\relax}@{}}
\toprule
\multicolumn{2}{@{}l}{\textbf{SieveVL-0.8B (Main)}}\\
\textbf{Stratum (pairs)} & \textbf{Pair / Brier / Cite F1 (Exact)} \\
\midrule
\texttt{zh/document\_region} (104) & 100.0\% / 0.0031 / 0.9927 (98.08\%) \\
\texttt{zh/table} (346) & 100.0\% / 0.0081 / 0.9485 (82.95\%) \\
\texttt{en/table} (52) & 100.0\% / 0.0152 / 0.9474 (80.77\%) \\
\texttt{zh/text} (45) & 95.56\% / 0.0440 / -- \\
\texttt{en/text} (36) & 100.0\% / 0.0165 / -- \\
\bottomrule
\end{tabular*}

\medskip
\begin{tabular*}{\linewidth}{@{\extracolsep{\fill}}>{\raggedright\arraybackslash}p{\dimexpr0.41\linewidth-4pt\relax}>{\raggedleft\arraybackslash}p{\dimexpr0.59\linewidth-4pt\relax}@{}}
\toprule
\multicolumn{2}{@{}l}{\textbf{SieveVL-0.8B (+KD)}}\\
\textbf{Stratum (pairs)} & \textbf{Pair / Brier / Cite F1 (Exact)} \\
\midrule
\texttt{zh/document\_region} (104) & 100.0\% / 0.0060 / 0.9856 (95.19\%) \\
\texttt{zh/table} (346) & 100.0\% / 0.0026 / 0.9501 (83.24\%) \\
\texttt{en/table} (52) & 100.0\% / 0.0106 / 0.9484 (82.69\%) \\
\texttt{zh/text} (45) & \textbf{100.0\%} / 0.0641 / -- \\
\texttt{en/text} (36) & 100.0\% / 0.0301 / -- \\
\bottomrule
\end{tabular*}

\medskip
\begin{tabular*}{\linewidth}{@{\extracolsep{\fill}}>{\raggedright\arraybackslash}p{\dimexpr0.41\linewidth-4pt\relax}>{\raggedleft\arraybackslash}p{\dimexpr0.59\linewidth-4pt\relax}@{}}
\toprule
\multicolumn{2}{@{}l}{\textbf{SieveVL-0.8B (Null-Cal)}}\\
\textbf{Stratum (pairs)} & \textbf{Pair / Brier / Cite F1 (Exact)} \\
\midrule
\texttt{zh/document\_region} (104) & \textbf{100.0\% / 0.0100 / 1.0000 (100.0\%)} \\
\texttt{zh/table} (346) & 100.0\% / 0.0050 / 0.9442 (80.35\%) \\
\texttt{en/table} (52) & \textbf{100.0\% / 0.0135 / 0.9569 (86.54\%)} \\
\texttt{zh/text} (45) & 97.78\% / 0.0674 / -- \\
\texttt{en/text} (36) & \textbf{100.0\% / 0.0008} / -- \\
\bottomrule
\end{tabular*}
\caption{Complete five-stratum internal validation results (583 paired examples, strictly source- and question-disjoint from training).}
\label{tab:supp_internal_val}
\end{table}

\begin{table}[tbp]
\centering
\footnotesize
\setlength{\tabcolsep}{2pt}
\begin{tabular*}{\linewidth}{@{\extracolsep{\fill}}>{\raggedright\arraybackslash}p{\dimexpr0.27\linewidth-4pt\relax}>{\raggedright\arraybackslash}p{\dimexpr0.73\linewidth-4pt\relax}@{}}
\toprule
\multicolumn{2}{@{}l}{\textbf{SieveVL-0.8B (Main, \texttt{r2\_listwise\_no\_kd})}}\\
\textbf{Setting / stratum} & \textbf{Value} \\
\midrule
Total groups / candidates & 9,430 groups / 23,892 filtered candidates (from 30,965 raw pool via 8B native hard-negative filtering to $\le 3$ distractors on image/text strata; KL weight $=0$) \\
Positive strata (groups) & \texttt{en/image} 4,351 (MMQA 2,500, WebQA 1,851); \texttt{zh/doc\_reg} 1,511; \texttt{zh/table} 1,500; \texttt{en/table} 804; \texttt{zh/text} 716; \texttt{en/text} 548 \\
Empty-pool groups & None \\
Trainable modules & LoRA ($r{=}16, \alpha{=}32$, drop $0.05$, 10.82M) + \texttt{visual.merger} (12.59M) + Heads (0.26M) \\
Learning rates & LoRA $1.5{\times}10^{-4}$, Merger $5{\times}10^{-5}$, Heads $2{\times}10^{-4}$ (5\% warmup) \\
Loss weights & $\mathcal{L}_{\mathrm{list}} + 0.30\mathcal{L}_{\mathrm{bce}} + 0.25\mathcal{L}_{\mathrm{cite}}$ ($\tau{=}1.0$) \\
Batching / Length / Steps & \texttt{max\_cands}=8, accum=8, \texttt{max\_tokens}=2,304, 395 steps (2,080.6\,s) \\
\bottomrule
\end{tabular*}
\medskip
\begin{tabular*}{\linewidth}{@{\extracolsep{\fill}}>{\raggedright\arraybackslash}p{\dimexpr0.27\linewidth-4pt\relax}>{\raggedright\arraybackslash}p{\dimexpr0.73\linewidth-4pt\relax}@{}}
\toprule
\multicolumn{2}{@{}l}{\textbf{SieveVL-0.8B (Null-Cal, \texttt{r4\_null\_calibrated})}}\\
\textbf{Setting / stratum} & \textbf{Value} \\
\midrule
Total groups / candidates & 12,930 groups (10,630 pos + 2,300 empty) / 42,365 cands \\
Positive strata (groups) & Full unfiltered 9,430-group (30,965-candidate) pool + 1,200 XFUND \texttt{zh.train} 6-strip positive groups (105 clean train-only pages) \\
Empty-pool groups & 800 MMQA native + 800 CLIP cross-Q (cos 0.2542) + 400 non-img + 300 strip \\
Trainable modules & Identical (23,675,394 trainable / 873,438,784 total) \\
Learning rates & Identical \\
Loss weights & $\mathcal{L}_{\mathrm{null}}(s_0{=}0) + 0.50\mathcal{L}_{\mathrm{bce}} + 0.35\mathcal{L}_{\mathrm{cite}}$ ($w_+ \le 3.0$) \\
Batching / Length / Steps & \texttt{max\_cands}=8, accum=8, \texttt{max\_tokens}=2,816, 813 steps (6,545.4\,s) \\
\bottomrule
\end{tabular*}
\caption{Exact training corpus manifest and optimization hyperparameters for SieveVL-0.8B (Main) and SieveVL-0.8B (Null-Cal). Both runs train for 1 epoch on a single NVIDIA A100 MIG \texttt{3g.40gb} slice with stripped image titles and verified zero page, document, or template overlap against all evaluation benchmarks.}
\label{tab:supp_hparams}
\end{table}

\paragraph{Exact Scoring and Reader Prompt Templates}
To ensure full reproducibility, all evaluations use deterministic prompts without task-specific post-hoc engineering:
\begin{itemize}
    \item \textbf{SieveVL-0.8B Scoring Prefix:} \emph{``Decide whether the current candidate provides useful evidence for the question. Answer only `Yes' or `No'.''} followed by \texttt{Question: \{q\}} and \texttt{Current candidate:} (raw image, page, table cells, or coordinate-indexed region payload).
    \item \textbf{Reproduced ACL 2026 Utility-VES~\cite{luo2026utility} Prompt:} Verbatim from \texttt{Hcnaeg/utility-mrag} (\emph{``Based on the images provided, does the retrieved image provide helpful visual or factual information that could assist in answering the question correctly? Answer with True or False.''} for MRAG-Bench; single-image variants for Visual-RAG and MP-DocVQA).
    \item \textbf{Decoupled 27B Reader Prompts (temperature 0.0, thinking disabled):} MRAG-Bench: \emph{``Choose the correct option for the query image. The retrieved examples, if any, may be irrelevant. Reply with only A, B, C, or D.''} MP-DocVQA: \emph{``Read the provided document page(s) and answer the question directly. Reply with ONLY the exact short answer string (no explanation or full sentence).''} XFUND / FUNSD: direct field-value extraction with explicit abstention instruction (\texttt{UNANSWERABLE}) when the cropped image does not contain the queried field, enabling causal verification under counterfactual cropping.
\end{itemize}
\FloatBarrier

\end{document}